\documentclass[journal]{IEEEtran}
\usepackage{xcolor,soul,framed} %,caption
\colorlet{shadecolor}{yellow}
\usepackage[pdftex]{graphicx}
\graphicspath{{../pdf/}{../jpeg/}}
\DeclareGraphicsExtensions{.pdf,.jpeg,.png}

\usepackage{array}
\usepackage{mdwmath}
\usepackage{mdwtab}
\usepackage{eqparbox}
\usepackage{url}
\usepackage{booktabs}
\usepackage{graphicx}
\usepackage{subfigure}
\usepackage{bm}
\usepackage{multirow}
\usepackage{float}

\let\labelindent\relax
\usepackage{enumitem}  %序号

\usepackage[cmex10]{amsmath}%公式分行

\usepackage{algpseudocode}  %伪代码
\usepackage{algorithm}  
\usepackage{amsmath}  

\usepackage{cite} %合并引用

\usepackage{hyperref}          %超链接
\hypersetup{hypertex=true,
	colorlinks=true,
	linkcolor=black,
	anchorcolor=black,
	citecolor=black}

\begin{document}
	\bstctlcite{IEEEexample:BSTcontrol}
	\title{RS-MetaNet: Deep meta metric learning  for few-shot remote sensing scene classification
	}
	
	\author{Haifeng Li,~\IEEEmembership{Member,~IEEE,}
		Zhenqi Cui,
		Zhiqing Zhu,
		Li Chen,
		Jiawei Zhu,
		Haozhe Huang,
		Chao Tao \thanks{}
		
		\thanks{This work was supported by the National Natural Science Foundation of China (grant numbers 41571397, 41871364, 41671357 and 41871302).}
		\thanks{H. Li, Z. Cui, Z. Zhu, L. Chen, J. Zhu, H. Huang, and C. Tao are with School of Geosciences and Info-Physics, Central South University (Corresponding author C. Tao, Email: kingtaochao@csu.edu.cn).}% <-this % stops a space
		
	}

	\markboth{Accept by IEEE TRANSACTIONS ON GEOSCIENCE AND REMOTE SENSING }{H. Li \MakeLowercase{\textit{et al.}}: RS-MetaNet}

	% ====================================================================
	\maketitle
	
	\setcounter{page}{1}
	
	% === ABSTRACT ====================================================================
	% =================================================================================
	\begin{abstract}
		Training a modern deep neural network on massive labelled samples is the main paradigm in solving the scene classification problem for remote sensing, but learning from only a few data points remains a challenge. Existing methods for few-shot remote sensing scene classification are performed in a sample-level manner, resulting in easy overfitting of learned features to individual samples and inadequate generalization of learned category segmentation surfaces. To solve this problem, learning should be organized at the task level rather than the sample level. Learning on tasks sampled from a task family can help tune learning algorithms to perform well on new tasks sampled in that family. Therefore, we propose a simple but effective method, called RS-MetaNet, to resolve the issues related to few-shot remote sensing scene classification in the real world. On the one hand, RS-MetaNet raises the level of learning from the sample to the task by organizing training in a meta way, and it learns to learn a metric space that can well classify remote sensing scenes from a series of tasks. We also propose a new loss function, called Balance Loss, which maximizes the generalization ability of the model to new samples by maximizing the distance between different categories, providing the scenes in different categories with better linear segmentation planes while ensuring model fit. The experimental results on three open and challenging remote sensing datasets, UCMerced\_LandUse, NWPU-RESISC45, and Aerial Image Data, demonstrate that our proposed RS-MetaNet method achieves state-of-the-art results in cases where there are only \bm{$1\sim20$} labelled samples.
	\end{abstract}

	% === KEYWORDS ====================================================================
	% =================================================================================
	\begin{IEEEkeywords}
		Remote sensing classification, Meta task, Metric learning, Few-shot learning
	\end{IEEEkeywords}

	% \fi
	\IEEEpeerreviewmaketitle

	% ====================================================================
	% ====================================================================
	% ====================================================================

	% === I. INTRODUCTION =============================================================
	% =================================================================================
	\section{Introduction}
	
	\IEEEPARstart {S}{cene} classification is an important component of optical remote sensing image processing and analysis and has wide applications in disaster detection, environmental monitoring, urban planning, land use and other national economic construction fields \cite{RN2,RN1,RN56,RN3,RN64}. Few-shot scene classification is achieved by using a very small number of labelled samples, which can effectively alleviate difficulties existing in traditional classification methods, such as open-set classification and domain difference between different datasets. According to the features used for remote sensing scene classification, existing approaches can be divided into methods based on handcrafted features and those based on deep features.
	
	The handcrafted features used for remote sensing scene classification can be grouped into three categories: spectral features, texture features, and structural features \cite{RN4}. From the literature in recent years, almost all methods based on handcrafted features, such as bag-of-visual-words (BOVW) models, probabilistic topic models (PTM) and sparse coding \cite{RN4,RN5,RN6}, encode these features before applying them to remote sensing scene classification to remove redundant information and increase the sparseness, rotation invariance and scale invariance of the features to improve classification performance. However, in practical applications, the performance is largely limited by handcrafted designed features, making it difficult to describe the rich semantic information contained in remote sensing images.
	
	In recent years, methods based on deep features have attracted more research attention\cite{RN10,RN12,RN57,RN14,RN58,RN59,RN61} due to the availability of large-scale training data and the development of high-performance computing units\cite{RN8}. The essence of such methods is that they extract features end to end with deep neural networks such as autoencoder, deep belief network\cite{RN9} and convolutional neural network (CNN). Earlier work generally extracted features or generated visual words from an off-the-shelf CNN\cite{RN62}, such as AlexNet\cite{RN10}, VGG16\cite{RN11}, and GoogleNet\cite{RN12}, and used an automatic encoder to implement remote sensing scene classification. Most recent work has generally increased the amount of information in the fused features by fusing the features of different layers of one or more CNNs to improve classification performance. For example, \cite{RN13} uses discriminant correlation analysis to fuse the features of two fully connected layers of VGG-Net\cite{RN11}. \cite{RN63} uses a shallow weighted deconvolution network to learn a set of feature maps and filters for each image by minimizing the reconstruction error between the input image and the convolution result. \cite{RN14} fuses the features of the convolutional layer and the fully connected layer of CaffeNet to obtain a new feature and then uses VGG-Net to perform the same operation to obtain another new feature. Finally, the linear combination method is used to combine the two new features. However, the success of these methods depends on large amounts of labelled data. These methods are ``data hungry’’ in nature because they learn each task independently from scratch by fitting a deep neural network over the data through extensive, incremental model updates using optimization algorithms such as stochastic gradient descent or Adam. Therefore, the existing approaches cannot learn a new data distribution well due to overfitting if the new remote sensing scene does not exist in the closed training dataset and has few labels. Therefore, the problems of fast adaptation in dynamic environments and limited data are fundamental challenges. For example, classical models, including Resnet\cite{RN15} and GoogleNet\cite{RN12}, can achieve at least 90\% classification accuracy on Aerial Image Data (AID), UCMerced\_LandUse (UCM)  and other datasets\cite{RN16,RN17} but less than 40\% accuracy when given only one labelled sample.

\begin{figure}
		\begin{center}
			\includegraphics[width=3.5in]{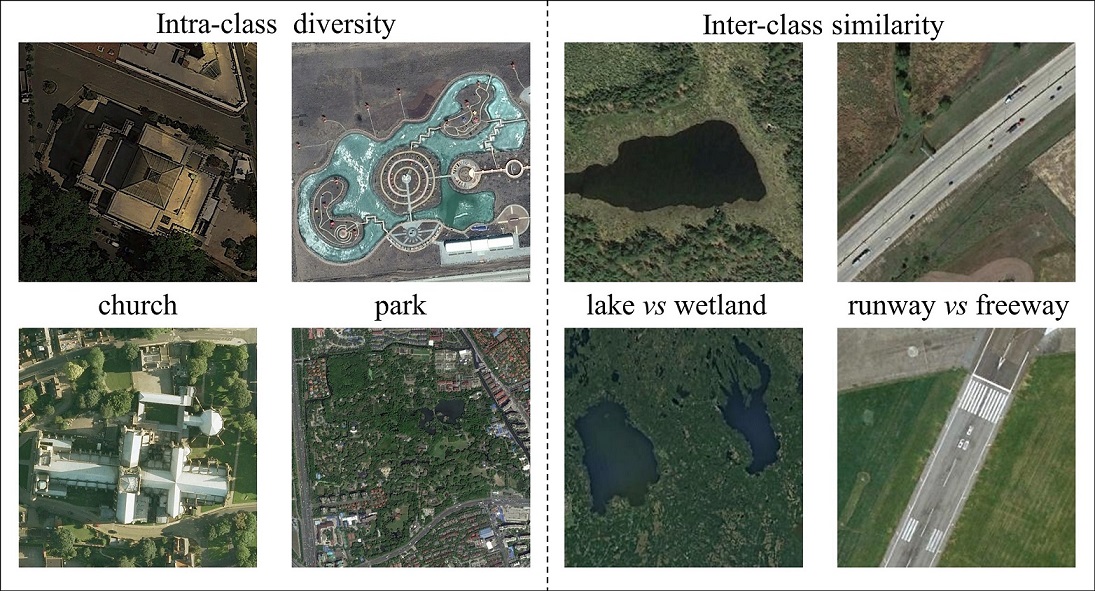}
			\caption{Different remote sensing images of church and park scenes are shown on the left, and remote sensing images of a lake and wetland and of a runway and freeway are shown on the right. An important challenge is to learn more distinctive feature representations with smaller intraclass dispersion but larger interclass separation. The above remote sensing images are taken from the public datasets AID and NWPU-RESISC45.}\label{fig:intra}
		\end{center}
	\end{figure}

	Therefore, increasing the level of learning and avoiding overfitting the data itself is a promising way to address few-shot remote sensing scene classification tasks. Meta-learning provides a new perspective on this problem\cite{RN19,RN18,RN20} by raising the level of learning from the data to the task, and it has been successful in classification\cite{RN21,RN22}, regression\cite{RN24,RN23}, and reinforcement learning tasks\cite{RN25} in machine learning. Meta-learning learns from a set of tasks rather than a set of data; each task consists of a labelled training set and a labelled testing set to simulate a few-shot learning problem, thus making the training problem more faithful to the real environment. This line of thinking has given us a great deal of inspiration, but remote sensing images are quite different from natural images due to their unique properties, such as the large variability in the same scene and similarity between different scenes\cite{RN26,RN54}, as shown in Fig.\ref{fig:intra} Therefore, another important challenge is to measure the similarity of tasks---in other words, to learn more distinctive feature representations with smaller intraclass dispersion but larger interclass separation.
	
	To explore a possible solution that addresses these challenges, we propose a framework called RS-MetaNet to improve the performance of few-shot remote sensing scene classification. On the one hand, our proposed RS-MetaNet trains the model through meta-tasks that simulate remote sensing scene classification tasks with very few samples. The meta-task forces our model to learn to learn task-based metrics, which learn a task-level distribution that should be better generalized to the unseen test task. On the other hand, we propose a new loss function called balance loss, which couples a maximum-generalization loss with the cross-entropy loss function used by traditional classification neural networks. Guided by the balance loss, RS-MetaNet maximizes the generalization ability of the model to new samples by maximizing the distance between different categories, providing the scenes in different categories with better linear segmentation planes while ensuring model fit. In addition, instead of explicitly defining distances, such as cosine distances or Euclidean distances, the distances in RS-MetaNet are learnable, which allows our method to fully utilize the intrinsic information in the data and thus make the metric space more distinct. We evaluate the proposed approach on three public benchmark datasets, and the experimental results show that our proposed RS-MetaNet method outperforms existing methods and achieves state-of-the-art results in cases where only 1-20 labelled samples are used.
	
	In summary, this work offers three contributions:
	
	\begin{itemize}[topsep = 0.5em,leftmargin = 1em]
		
		\item {We propose a novel framework, called RS-MetaNet, to improve the performance of few-shot remote sensing scene classification. RS-MetaNet trains the model through meta-tasks, which forces our model to learn to learn a task-level distribution that should be better generalized to the unseen test task.}

		\item{We propose a new loss function, called balance loss, which couples a maximum-generalization loss with the cross-entropy loss function. Within the constraints of the loss, RS-MetaNet maximizes the generalization ability of the model to new samples by maximizing the distance between different categories, providing the scenes in different classes with better linear segmentation planes while ensuring model fit.}
		
		\item{Our metric module, using learnable distance metrics, allows RS-MetaNet to take full advantage of the information in the data itself, making the learned metric space more discriminating.
		}
		
	\end{itemize}

	The rest of our paper is organized as follows: In Section \ref{Related Work}, we introduce works related to few-shot remote sensing scene classification. Section \ref{sec:Proposed Method} describes the proposed method. We describe the experiments in Section \ref{sec:Experiment and analysis}, and  finally, discussions and conclusions are presented in Section \ref{sec:Discussion} and \ref{sec:conclusion} respectively.

	\section{Related Work}\label{Related Work}
	
	% =======
	% FIG. 01
	% =======
	
	% =======
	% FIG. 02
	% =======
	\begin{figure*}
		\begin{center}
			\includegraphics[width=7in]{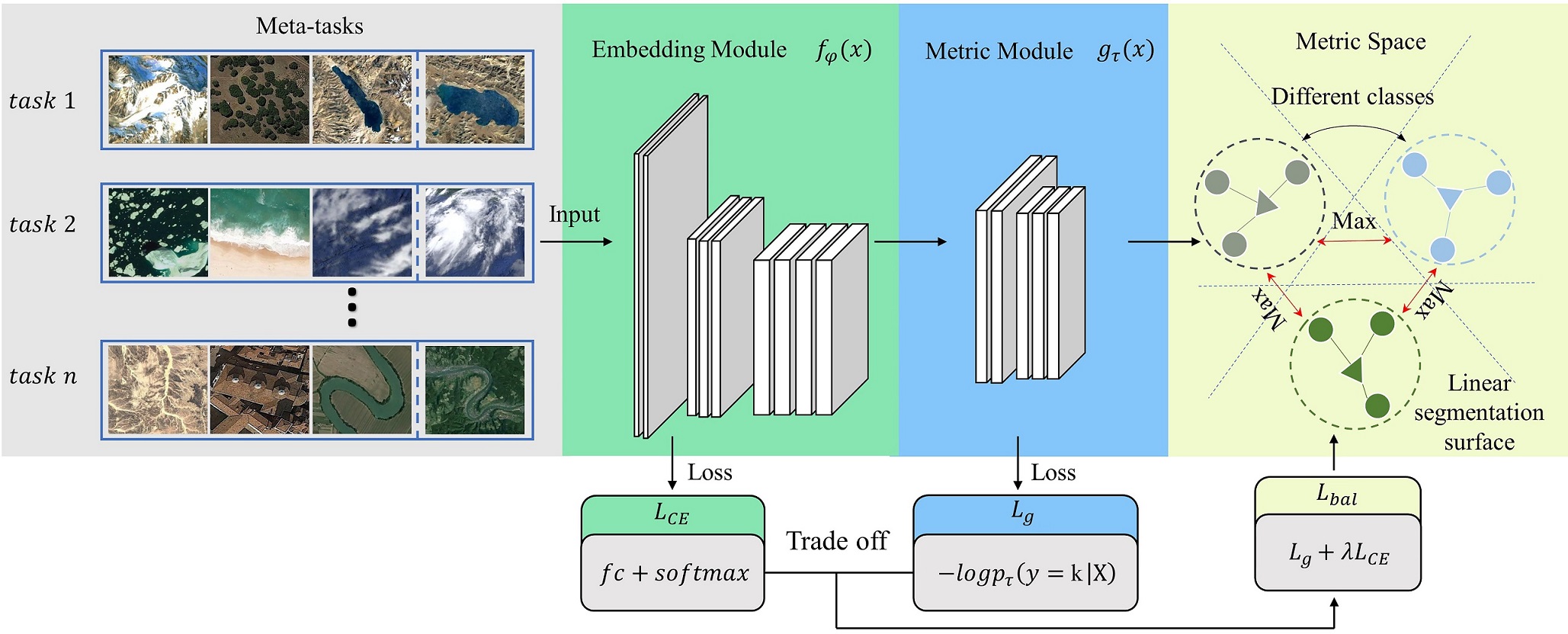}
			\caption{\textbf{Overview.} Schematic illustration of our method RS-MetaNet on three-class classification tasks of remote sensing. RS-MetaNet learns to learn a metric space that can maximize the distance between different classes through meta tasks so that there are better segmentation surfaces between them. Thus, the task-based metric learns a task-level distribution that should be better generalized to the unseen test task (see Section \ref{sec:Proposed Method} for details).}\label{overview}
		\end{center}
	\end{figure*}

	There are few remote sensing scenes on a global scale, and it is difficult to obtain a large number of samples at a wide range of temporal and spatial scales, which severely limits the scalability of existing scene classification methods to new remote sensing  scene categories and to rare categories that may lack many labelled images. However, at present, only a handful of methods have attempted to solve few-shot scene classification in the field of remote sensing\cite{RN46,RN41,RN55}.

	\subsection{Metric Learning}\label{sec:Metric Learning}
	
	The metric learning problem is concerned with learning a distance function tuned to a particular task, and its essential idea is to make the samples of the same category as close as possible in the custom space and keep the samples of different categories as far apart as possible\cite{RN28,RN29,RN27}. Solving this problem has shown to be useful for remote sensing technology. \cite{RN26} applies metric learning regularization terminology to the function of CNNs to make their proposed discriminative CNN model more discriminative. \cite{RN30} selects the nearest neighbour algorithm combined with metric learning to perform similarity learning between paired samples. \cite{RN31} further improves the model representation ability by increasing the variance between different centre points and reducing the variance between the features learned from each class and the corresponding centre points. Similarly, \cite{RN32} applies metric learning to hyperspectral target detection. This method targets the information-theory metric learning method and is used to learn a Mahalanobis distance so that similar and dissimilar point pairs can be separated without similarity assumptions . \cite{RN33} applies a triangular loss function to consider the whole relation within a tuple and successfully improved the effect of remote sensing image retrieval.
	
	\subsection{Transfer Learning}\label{sec:Transfer Learning}
	
	Transfer learning aims to use knowledge of the source domain to improve or optimize the learning effect of the target prediction function in the target domain\cite{RN34,RN35}. For deep learning models, adjusting the pre-trained model for new tasks, which is often called fine-tuning, is a powerful transfer method. It has been shown that pre-trained models on large-scale datasets are better generalized than randomly initialized models\cite{RN36}. \cite{RN37} distinguishes images of 1000 categories well by retaining the parameters of all convolutional layers in a trained Inception-v3\cite{RN38} model and simply replacing the last fully connected layer. \cite{RN39} indicates that deciding what and how to transfer are the key steps in transfer learning because different methods are applied to different source-target domains and bridge different transfer knowledge domains\cite{RN40}. \cite{RN41} trained two deep encoders to transfer knowledge in the Electro-Optica domain by learning a shared invariant cross-domain embedding space; the Sliced Wasserstein distance was used to measure and minimize the distance between domains, and a limited number of labelled data points were used to match the distributions in a class-conditional manner.
	
	\subsection{Meta-learning}\label{sec:Meta-Learning}
	
	Meta-learning is currently a popular topic in the machine learning community\cite{RN22,RN23,RN42}. Meta-learning, or learning to learn, is the science of systematically observing how different machine learning methods perform in a wide range of learning tasks and then learning from this experience or meta-data to learn new tasks faster than otherwise\cite{RN43}. Meta-learning focuses more on tasks than data. In meta-learning, a task refers to a separate complete classification of tasks, and each task consists of two parts, meta-training and meta-testing, corresponding to the training and testing processes in traditional deep learning. \cite{RN21} use LSTM-based meta-learners to learn update rules for training neural network learners. \cite{RN23} learns a model parameter initialization that generalizes well to similar tasks. \cite{RN44} notes that the MAML algorithm can be understood as inferring the parameters of the prior distribution in the Bayesian model; REPTILE \cite{RN45} is an approximation of MAML, which performs K iterations of stochastic gradient descent for a given task and then gradually moves the initialization weights towards the weights obtained after K iterations. Similarly, \cite{RN46} also learns model parameter initialization parameters. It iteratively selects a batch of previous tasks, trains learners for each task to calculate gradients and losses, and updates in a direction in which the weights are more easily updated by back-propagation.

	\section{Proposed Method}\label{sec:Proposed Method}
	
	This work focuses on addressing two challenges in remote sensing scene classification in the real world: 1) the trained model must operate on new remote sensing scenes that did not appear in the closed training set; and 2) the trained model must be effective for unseen remote sensing scenes with only a few labelled samples. %QCE: Please ensure that the intended meaning has been maintained in the above edit. 
	An overview of our RS-MetaNet is presented in Fig.\ref{overview}.

	\subsection{Meta Training}
	
\begin{figure}
		\begin{center}
			\includegraphics[width=3.3in]{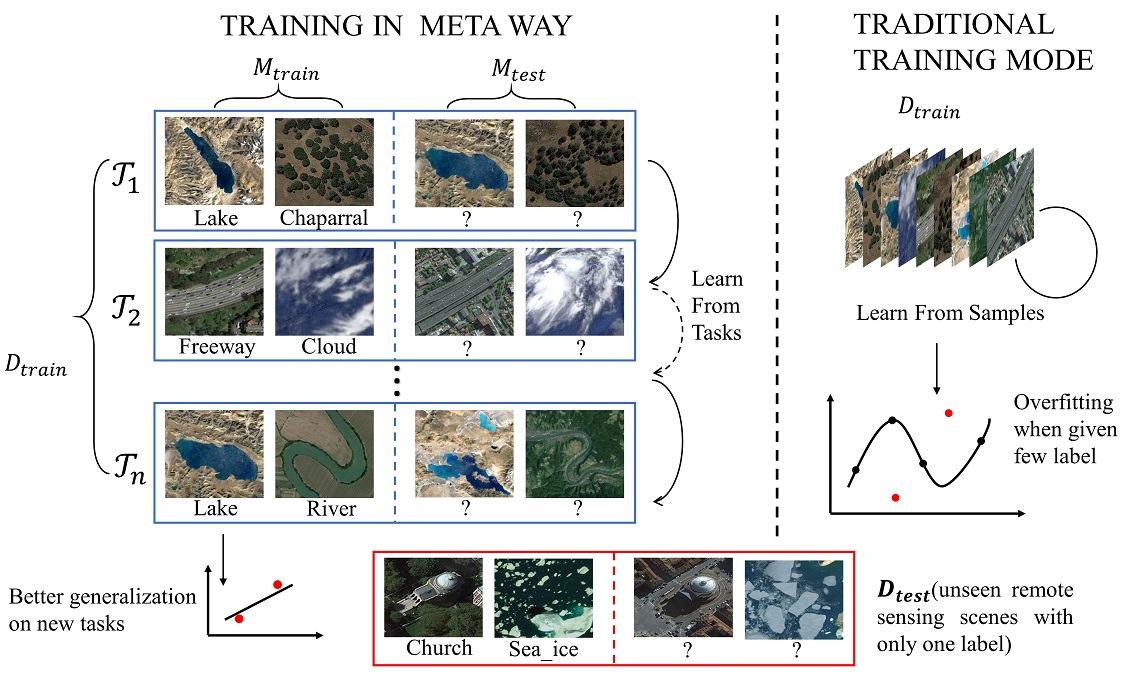}
			\caption{\textbf{Meta training.} Schematic illustration of binary few-shot remote sensing scene classification. For each task $T_i$, we use $M_{train}$ to learn the metric space and then use $M_{test}$ to update it. This allows our model to learn the task-based metric space and make it generalize well to unseen scenes in $D_{test}$.}\label{meta-way}
		\end{center}
	\end{figure}

	Traditional models essentially fit a deep model over the data and update the parameters step by step following the opposite direction of the gradient of the loss to learn the optimal model based on the data. However, these approaches ignore information between tasks and therefore restrict the scalability of the model to new tasks. Unlike the standard supervised training paradigm, we train our models in a meta training way, as shown in Fig.\ref{meta-way}. We learn a metric space on different tasks sampled from a task family that is tuned to perform well on new tasks sampled from this family.
	
	Specifically, given two non-overlapping sets of remote sensing scenes $C_{seen}$ and $C_{unseen}$, the training set $D_{train}$ is constructed from $C_{seen}$, and the testing set
	$D_{test}$ is constructed from $C_{unseen}$. Both the training set $D_{train}$ and testing set $D_{test}$ consist of a meta-training set $M_{train}$ and a meta-testing set $M_{test}$. Namely, $D_{train} = \left\{M_{train},M_{test}\right\}_{i=1}^{N}$ and
	$D_{test} = \left\{M_{train},M_{test}\right\}_{i=1}^{M}$, where N and M denote the
	numbers of tasks for training and testing, respectively. In this way, we enable each task to simulate few-shot remote sensing image classification, which makes our RS-MetaNet generalize well to the unseen test task. In other words, our model learns the ability to perform few-shot learning of classification instead of learning task-specific model parameters. 
	
	During the training phase, we build the meta-training set $M_{train}=\left\{(x_i,y_i)\right\}_{i=1}^{CS_{tr}}$
	by sampling $C$ different classes from $D_{train}$, and $S_{tr}$ labelled samples from each class; correspondingly, we build $M_{test}=\left\{(x_i,y_i)\right\}_{i=1}^{QS_{te}}$ by sampling $Q$ different classes from $D_{train}$ and $S_{te}$ labelled samples from each class, where the classes in $M_{test}$ are a proper subset of the classes in $M_{train}$. It is generally assumed that the training set $D_{train}$ and the testing set $D_{test}$ are sampled from the same distribution. To optimize the generalization error, it is stipulated that the meta-training set and meta-testing set in each task cannot have overlapping parts; that is, $M_{train} \cap M_{test} = \emptyset$. Moreover, the validation set $D_{val}$ separated from $C_{seen}$, which is designed to select the learner's hyperparameters and to choose the best model, does not intersect with $D_{train}$ or $D_{test}$. 
	
	%\begin{align}\label{1}
	%    \omega&=\mathop{\arg\min}_\omega %\sum\limits_{i=1}^{M}L_{bal}(D_{test};\omega,\theta(\omega)) \\
	%    s.t.\quad\theta(\omega)&=\mathop{\arg\min}_\theta L_{bal}(D_{train};\varphi,\tau)
	%\end{align}

	% =========
	% FIG. 03
	% =========
	
	\subsection{Embedding Module}
	
	In this module, we aim to use the embedding model $f_\varphi(x)$ parameterized by $\varphi$ to map the data domain to the feature space so that the visual information can be related to each other. The feature representation $V$ in the embedding space can be expressed by
	
	\begin{equation}\label{feature}
	V = f_\varphi(M_{train};\varphi)
	\end{equation}
	
	On the one hand, in each task, the embedding model $f_\varphi(x)$ should minimize the fitting loss $L_{CE}$ on the meta-training set $M_{test}$, which is defined as
	
	\begin{equation}\label{L_CE}
	L_{CE} = -\sum\limits_{i=1}^{CS_{tr}}(y_i\log{\hat{y}_i}  + (1-y_i)\log{(1-\hat{y}_i)}
	\end{equation}

	The fitting loss $L_{CE}$ intuitively represents the quality of the feature representation $V$ in the embedding space, which plays an important role in the final classification accuracy. Formally, for $M_{train}=\left\{(x_i,y_i)\right\}_{i=1}^{CS_{tr}}$, the learning object of the embedded module is defined by

	\begin{equation}\label{theta}
	\varphi=\mathop{\min}_\varphi\ E_{CS_{tr}}[L_{CE}(M_{train};\varphi)]
	\end{equation}
	
	On the other hand, since the final classification accuracy is affected by the quality of the feature space, we try to reduce the dimensional loss of $V$ 
	in Eq.(\ref{feature}); however, a higher dimensionality will bring a greater calculation burden for the subsequent metric module. Therefore, we use discriminative centroids for class structure replacement, inspired by\cite{RN48,Rn47} to reduce the computational complexity, which is computed by
	
	\begin{equation}\label{center}
	O_k = \frac{C}{|{M_{train}}|}\sum\nolimits_{(x_i,y_i)\in M_{train}^k}V_{x_i}
	\end{equation}

	where $M_{train}^k$ denotes the set of data labelled with class $k$ in $M_{train}$. This operation can help us increase the speed of calculation considerably by sacrificing a small amount of accuracy compared to using the representation of the embedded space to perform the operation directly.
	
	\subsection	{Metric Module}
	
	\begin{algorithm}[t] 
		\caption{ RS-MetaNet Algorithm}  
		\label{alg:Framwork}  
		\begin{algorithmic}[1]  
			\Require  
			Training remote sensing scene image set $D_{train}$, the number of classes in $M_{train}$ $C$, the number of samples per class  in $M_{train}$  $S_{tr}$, the number of samples per class in $M_{test}$ $S_{te}$, hyperparameter $\lambda$ , the number of tasks $T$.  
			\Ensure  
			Parameters $\varphi$ of the embedding model $f_\varphi(x)$; 
			Parameters $\tau$ of the metric model $g_\tau(x)$;
			\State Initialize $\varphi$ and $\tau$;  
			\For{$task = 1, 2, \ldots , T$}:
			\State Randomly sample $C$ class from $D_{train}$;  
			\State Randomly sample $S_{tr}$ instances for each class to build \quad$M_{train}$;
			\State Randomly sample $S_{te}$ instances for each class to build\quad $M_{test}$;
			\State Compute feature representation $V$ using (\ref{feature}): $V = f_\varphi(M_{train};\varphi)$;
			\State Metric space quadratic mapping using (\ref{center}) and (\ref{metric});  
			\State Optimize $\varphi$, $\tau$ using (\ref{L_BAL});  
			\EndFor 
			\State\Return $\varphi$,$\tau$;
		\end{algorithmic}  
	\end{algorithm}
	
	The metric module is one of the most important modules in our proposed RS-MetaNet model. We aim to maximize the generalization ability of the model by passing it to learn a metric space that can maximize the distance between different categories using a few feature representations, or even one, obtained by Eq.(\ref{feature}) and (\ref{center}). Specifically, for the obtained feature representation $V$, our goal is to learn a metric rule using the metric model $g_\tau$, parameterized by $\tau$, to maximize the discriminative ability of the metric space. $g_\tau$ consists of a single-layer neural network and a nonlinear activation function $ReLU(x)$\cite{RN49}, which is defined as
	
	\begin{equation}\label{relu}
	ReLU(x) =
	\begin{cases}
	x, & if\quad v > 0\\
	0, & if\quad v \leq 0
	\end{cases}
	\end{equation}

	To address the problem that samples in the new scene are too sparse, rather than explicitly defining the distance between samples, $g_\tau$ learns a metric rule that can maximize the distance among different categories of feature representations $V$ in the space; this rule  can make full use of the intrinsic information in the data. Specifically, for a point $X \in M_{test}$, we must optimize the parameter $g_\tau$ to maximize the distance between different categories based on a softmax over distances to the centroids in the embedding space in each task, which can be expressed as

	\begin{equation}\begin{split}\label{metric}
	\tau & = \mathop{\arg\max}p_\tau(y =k|X) \\
	&=\frac{\exp{(-g_\tau(f_\varphi(X),O_k)}}{\sum\nolimits_{O_k'}\exp{(-g_\tau(f_\varphi(X),O_{k'}))}},where\hskip 0.5pc k\in N
	\end{split}\end{equation}
	
	\subsection {Loss Function}
	
	Our proposed RS-MetaNet can be trained end to end by updating each module in turn since all our modules are differentiable. In contrast to Eq.(\ref{L_CE}), the metric model focuses more on the generalization ability of the model than the fitting ability. For point $X \in M_{test}$, the generalization loss $L_g$ is defined as
	
	\begin{equation}\label{L_G}
	L_g=-\log{p_\tau(y=k|X)}
	\end{equation}
	
	By the trade-off between fitness and generalization, we define the balance loss function $L_{bal}$ as follows:
	
	\begin{equation}\label{L_BAL}
	L_{bal}=L_g+\lambda L_{CE}
	\end{equation}
	
	$l_{CE}$ is defined in Eq.(\ref{L_CE}). $\lambda \in [0,1]$ is a hyperparameter that characterizes the tendency of the model. The smaller $\lambda$ is, the stronger the generalization ability of the model, and the larger $\lambda$ is, the stronger the fitting ability of the model. Since our model is designed for new remote sensing scenes that did not appear in the closed dataset and have few labelled data points, we need a principled approach that maximizes the generalization power rather than data fitting, so we constrain only the fitting error $l_{CE}$ and not the generalization error $L_g$.
	
	For a clearer explanation, we provide Algorithm \ref{alg:Framwork}, which details the RS-MetaNet procedure.
	
	\section{Experiment and Analysis}\label{sec:Experiment and analysis}
	
	In this section, we evaluate the proposed RS-MetaNet method on three public datasets designed for remote sensing scene classification. We first introduce the dataset used in the experiment (subsection \ref{sec:Data Set Description}). We then describe the network architecture and hyperparameter design in detail (subsection \ref{sec:Network Architecture}). Finally, we analyse our model and present our experimental results on different datasets (subsection \ref{sec:Classification Accuracy}$\sim$\ref{sec:Hyperparameter Analysis}).

	\subsection{Dataset Description}\label{sec:Data Set Description}
	
	In the experiments, we evaluate the proposed RS-MetaNet method on three publicly available datasets designed for
	remote sensing image scene classification.
	
	\textbf{The UCMerced\_LandUse dataset}\cite{RN17} contains 21 kinds of land use categories: agricultural, airplane, baseball diamond, beach, buildings, chaparral, dense residential, forest, freeway,
	golf course, harbour, intersection, medium density residential, mobile home park, overpass, parking lot, river, runway, sparse residential, storage tanks, and tennis courts. For each scene category, there are 100 aerial images in the red-green-blue (RGB) colour space, with a size of $256\times256$ pixels and a spatial resolution of $0.3m$. Since its emergence, this dataset has been widely used for remote sensing scene classification.

	\textbf{AID}\cite{RN16} is a large-scale dataset used for aerial scene classification including agricultural, airplane, baseball diamond, beach, buildings, chaparral, dense residential, forest, freeway, golf course, harbour, intersection, medium density residential, mobile home park, overpass, parking lot, river, runway, sparse residential, storage tanks, and tennis courts. It contains a total of 10,000 images in 30 scene categories with a size of $600\times600$ pixels. The number of images varies from 220 to 420 under different aerial scene categories. The spatial resolution varies from approximately 0.5 m to 8 m.
	
	% =========
	% Fig. 04
	% =========
	\begin{figure*}[!t]
		\begin{center}
			\includegraphics[width=7in]{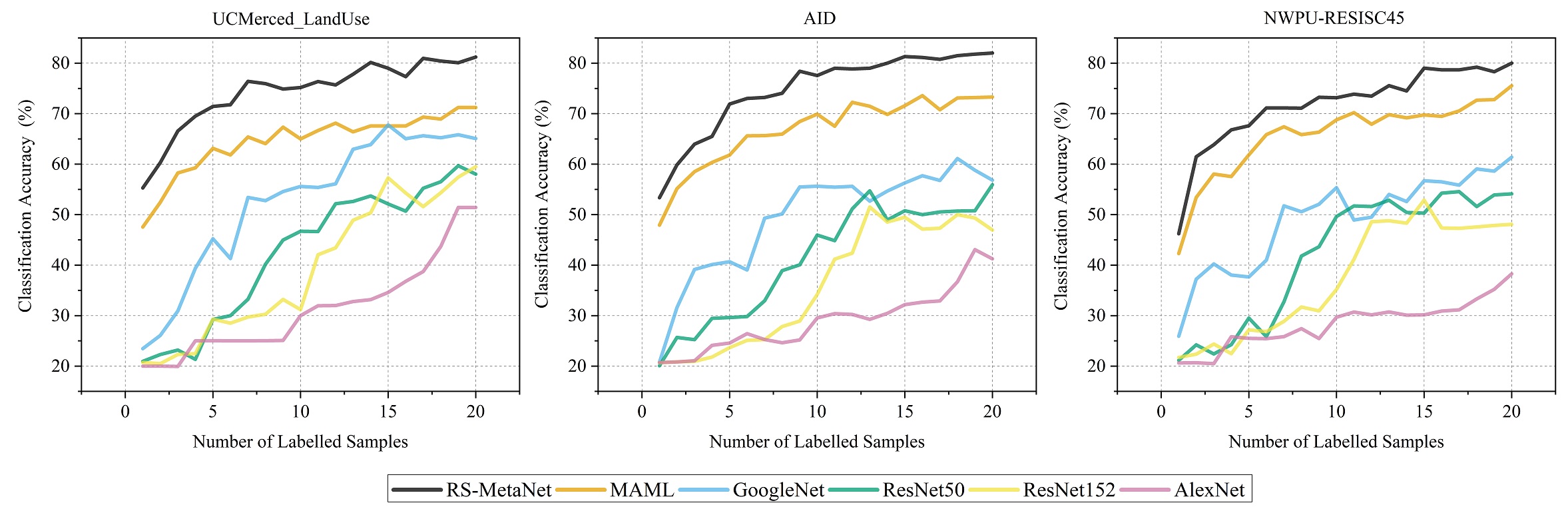}
			\caption{\textbf{Classification accuracy on three datasets.} The horizontal axis represents the number of labelled samples given on $D_{test}$. Our proposed RS-MetaNet significantly outperformed the baseline, especially when only one labelled sample was used.}\label{acc}
		\end{center}
	\end{figure*}

	\textbf{The NWPU-RESISC45 dataset}\cite{RN50} contains a total of 31,500 images divided into 45 scene categories: airplane, airport, baseball diamond, basketball court, beach, bridge, chaparral, church, circular farmland, cloud, commercial area, dense residential, desert, forest, freeway, golf course, ground track field, harbour, industrial area, intersection, island, lake, meadow, medium residential, mobile home park, mountain, overpass, palace, parking lot, railway, railway station, rectangular farmland, river, roundabout, runway, sea ice, ship, snowberg, sparse residential, stadium, storage tank, tennis court, terrace, thermal power station, and wetland. %QCE: Please consider replacing “snowberg” with “iceberg” if the intended meaning is maintained. 
	Each category consists of 700 images in the RGB colour space and is $256\times256$  pixels in size. For most scene categories, the spatial resolution varies from approximately 0.2 m to 30 m per pixel. To the best of our knowledge, this dataset is the largest in terms of the number of scene categories and the total number of images, with rich image variation, large intra-category diversity and high inter-category similarity, making the dataset more challenging.

	In the experiment, the remote sensing scenes in each dataset are evenly divided into three splits. Our model is trained on two splits and evaluated on the remaining one in a cross-validation fashion. For each test task, we randomly sample five scenes from  $D_{test}$ to simulate five new remote sensing scenes in the real world. Each scene is assigned only one or a few labelled samples for scene classification tasks. For the sake of fairness, the illustrated data divisions of the above three datasets are randomly generated.
	
	\subsection{Network Architecture and Hyperparameter Design}\label{sec:Network Architecture}
	
	For fairness, our initial embedding model has four modules with $3\times3$ convolutions and 64 filters, with padding = 1 to retain as much feature information as possible, followed by batch normalization\cite{RN51}, ReLU nonlinearity\cite{RN49}, and $2\times2$ max-pooling. How the embedded model affects the experimental results is shown in section \ref{sec:Effect Study}. We did not perform image enhancement processing. During the training phase, the weight decays every 20 epochs, and the corresponding parameter is set to 0.0005. The learning rate is set to 0.001. The metric module consists of a layer of a convolutional neural network and a fully connected layer. For the balance loss function, $\lambda$ is set to 0.1. We discuss the impression of parameter $\lambda$ on the model in detail later.

	\subsection{Classification Accuracy}\label{sec:Classification Accuracy}
	
	We compare the proposed RS-MetaNet method with the transfer learning-based method and meta-learning-based method MAML, which is popular in the meta-learning community. For transfer learning, we individually use GoogleNet, ResNet50, ResNet152, and AlexNet\cite{RN12,RN15,RN10} as the backbone, and they are pre-trained in advance on a large dataset. After training on $D_{train}$, once the accuracy of the model reaches at least 95\%, we randomly sample $1\sim20$ labelled samples of five classes in $D_{test}$ to fine-tune the model. Finally, the model is tested through classification tasks by the remaining samples of these classes, which has not been performed previously. %QCE: Please ensure that the intended meaning has been maintained in the above edit. 
	RS-MetaNet is trained from scratch on the training set $D_{train}$ without the need for pre-training or fine-tuning, and MAML needs fine-tuning but does not require pre-training.
	
	We evaluate experimental results using classification accuracy (acc), which is defined as
	\begin{equation}\label{eq:acc}
	Acc = \frac{1}{M}\sum\limits_{i=1}^{M}{\frac{r^{(i)}}{QS_{te}}}
	\end{equation}
	
	where $r^{(i)}$ denotes the number of correctly classified samples in task i.

\begin{figure}[t]
		\begin{center}

			\subfigure[One lable On UCM]{
				\includegraphics[width=3.8cm]{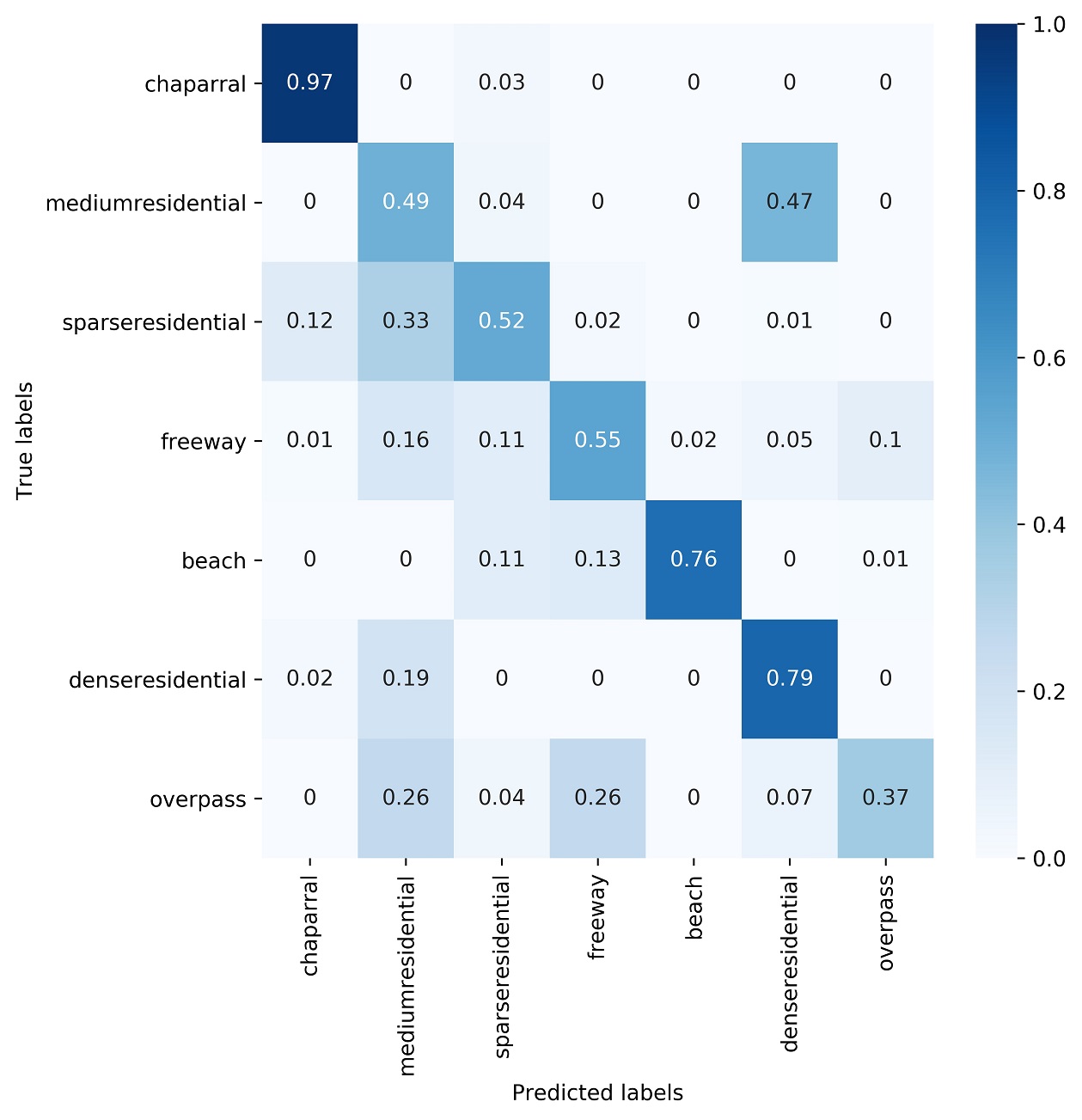}
			}
			\quad
			\subfigure[Five Lables On UCM]{
				\includegraphics[width=3.8cm]{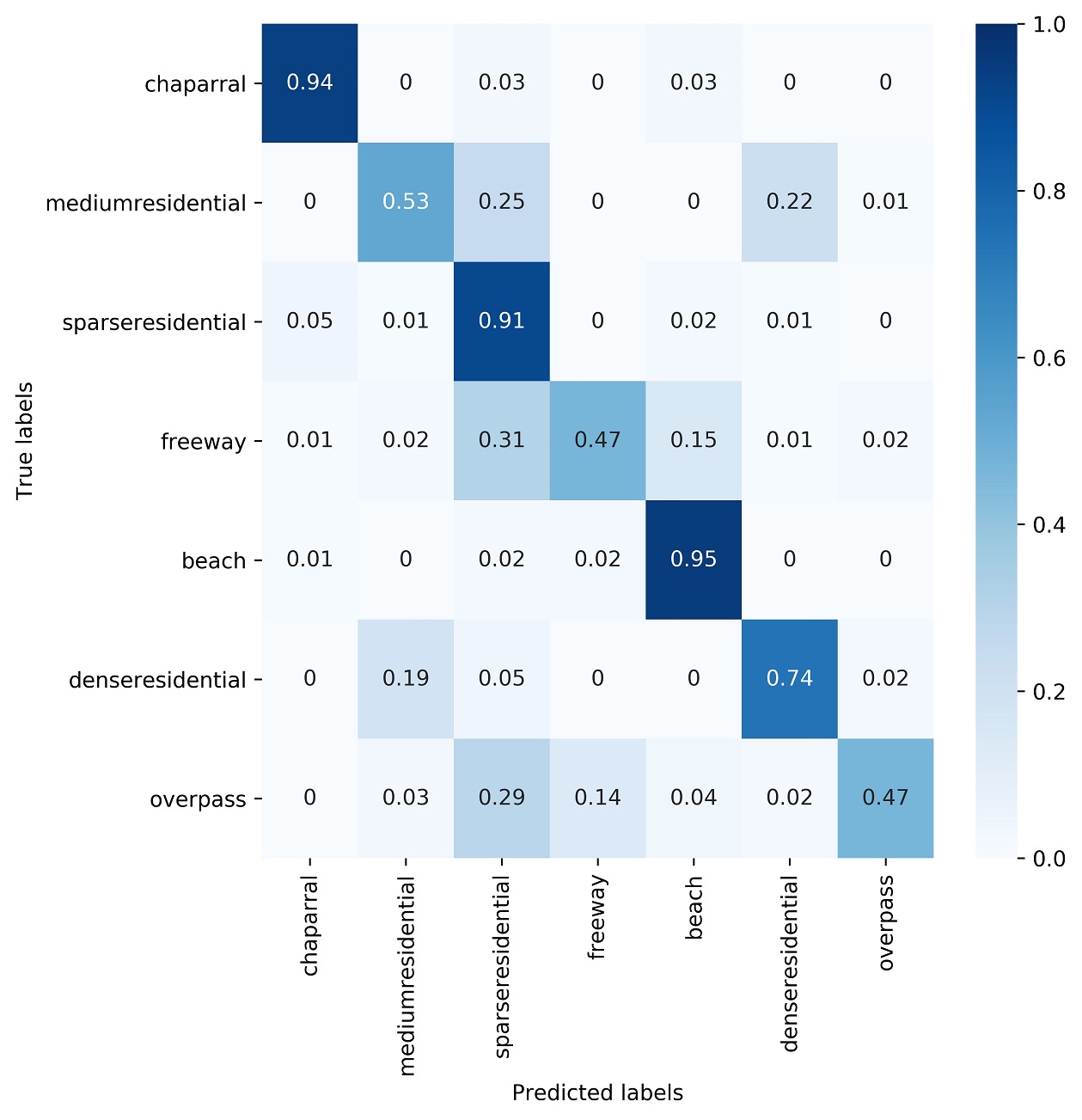}
			}
			\quad
			\subfigure[One Lable On AID]{
				\includegraphics[width=3.8cm]{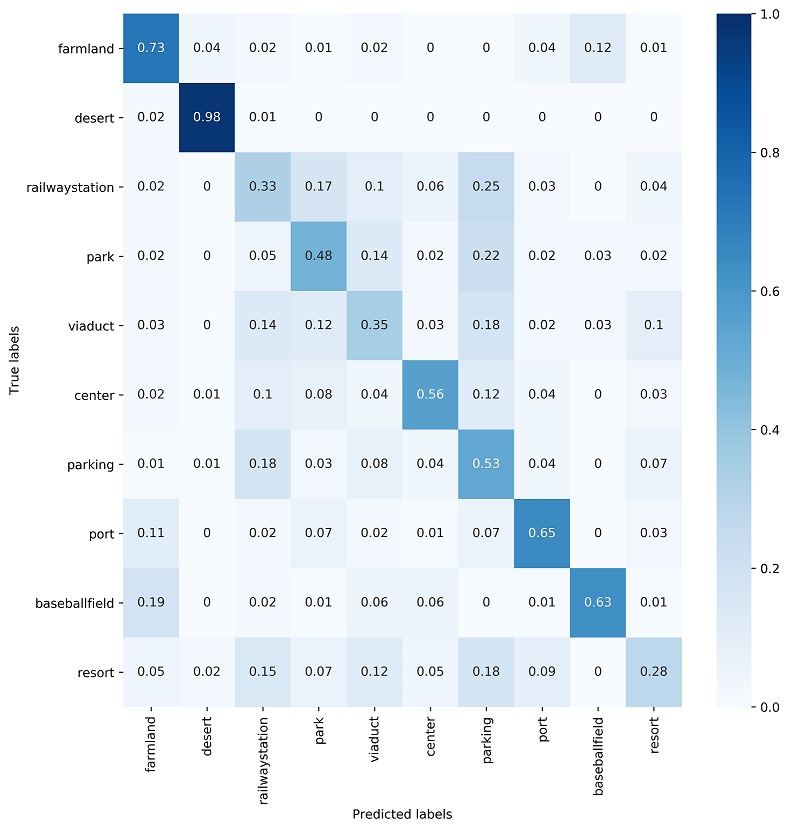}
			}
			\quad
			\subfigure[Five Lables On AID]{
				\includegraphics[width=3.8cm]{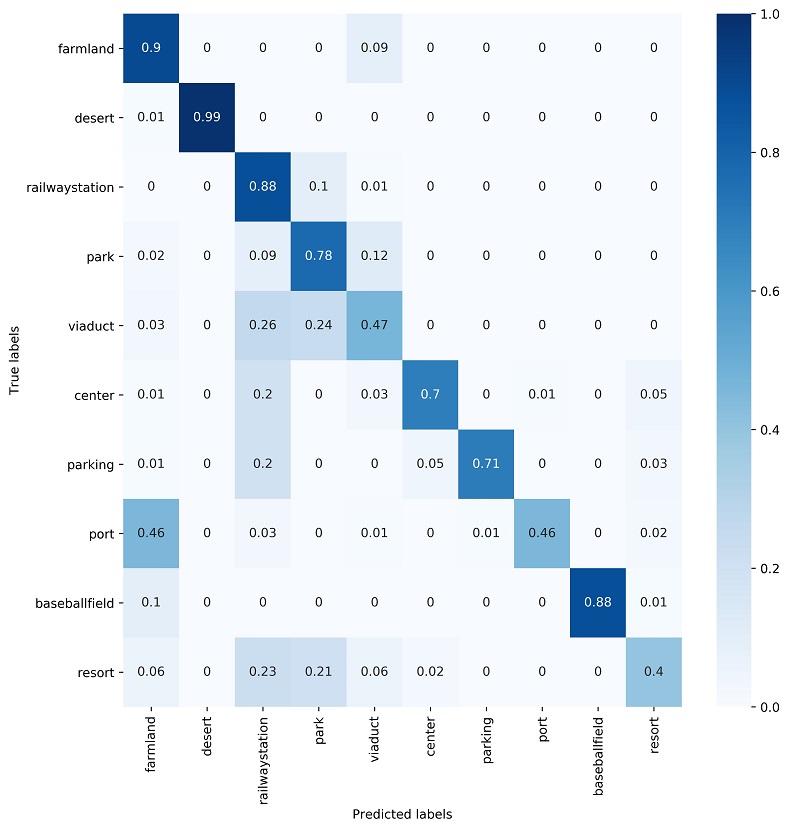}
			}
			\quad
			\subfigure[One lable On NWPU]{
				\includegraphics[width=3.8cm]{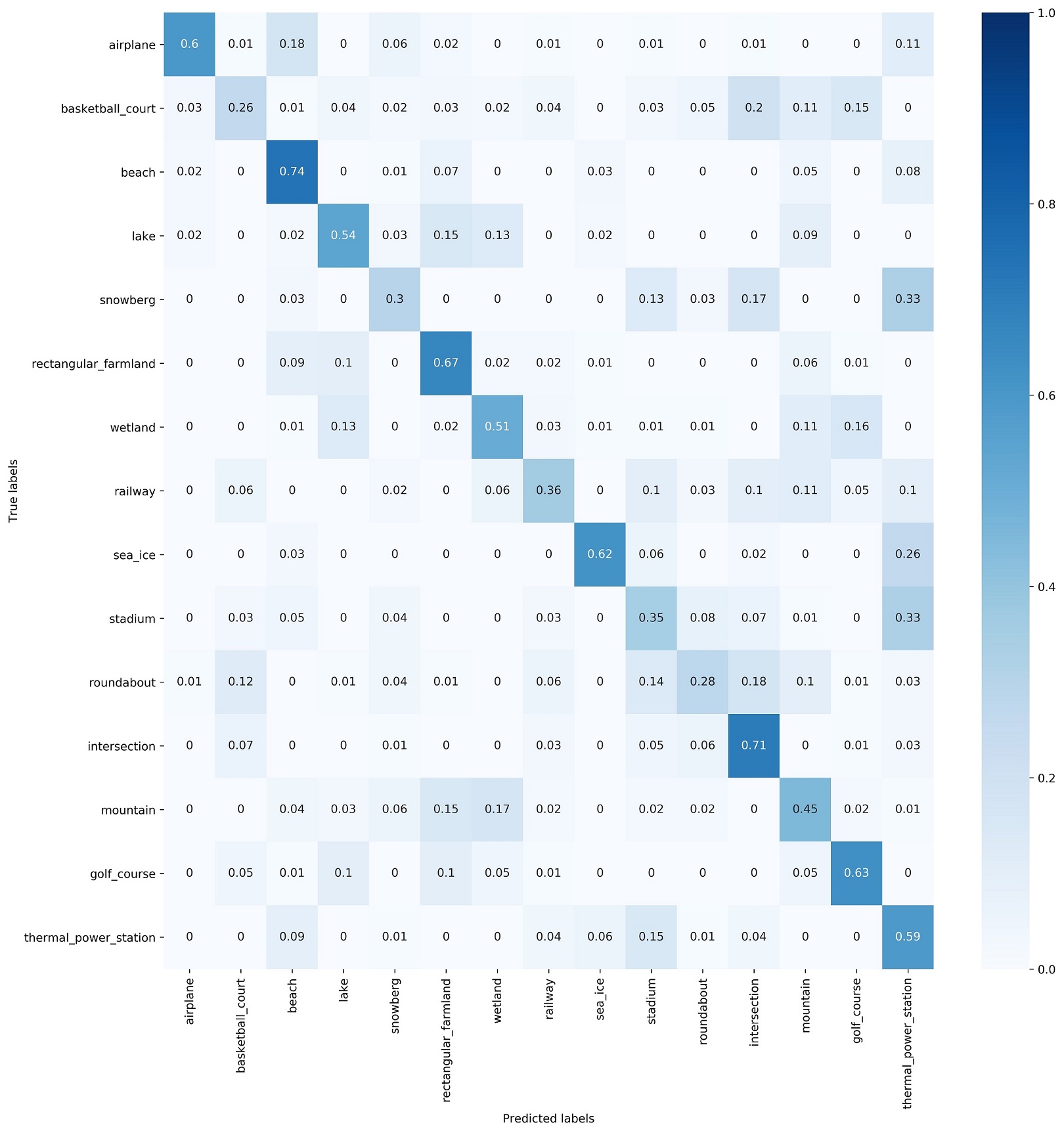}
			}
			\quad
			\subfigure[Five Lables On NWPU]{
				\includegraphics[width=3.8cm]{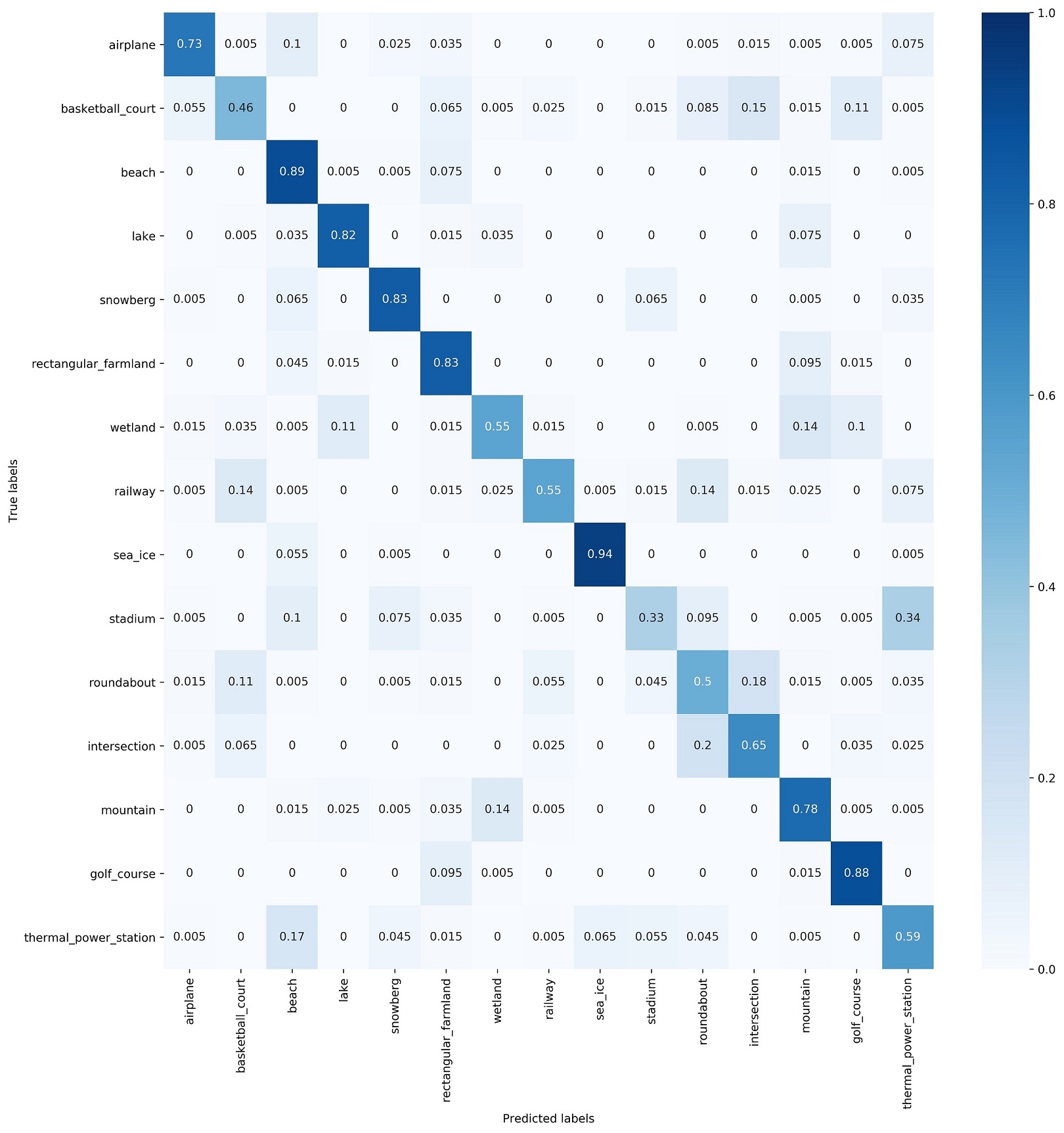}
			}
			\caption{\textbf{Confusion matrices} of the UCMerced\_LandUse dataset (top), AID (middle), and NWPU-RESISC45 dataset (bottom). The result with only one label is shown on the left, and that with five labels is shown on the right.}\label{Confusion matrices} 
			
		\end{center}
	\end{figure}

	All experiments in this paper are set as five-class classification. In theory, the smaller the number of classification scenes is, the higher the classification accuracy. It is difficult to use all samples in $D_{test}$ to fine-tune the model because only one sample is used for most test processes (otherwise, hundreds of experiments may be needed). Therefore, we perform 20 experiments by sampling at random from $D_{test}$ to eliminate the model's preference for data, and then we average the results of 20 experiments. As shown in Fig.\ref{acc}, our method is significantly better than MAML and transfer-learning-based methods, with a smaller variance. Especially when there are only one or two labelled samples, our proposed RS-MetaNet shows outstanding results. As the number of labelled samples increases, the model performance also gradually increases. Since our purpose is to perform few-shot remote sensing scene classification, the maximum number of given labels is set to 20. Additionally, we report the results measured in a confusion matrix. Due to space limitations, we present only the confusion matrix for three datasets given one and five labelled samples, as shown in Fig.\ref{Confusion matrices}, where the entry in the i-th row and j-th column represents the rate of test images from the i-th class that are classified as the j-th class.
	
	In addition, we consider that few-shot classification actually behaves in two ways, one with few known categories and the other with few scenes per category, so we analyse the performance of the model for different ratio training sets. Specifically, we randomly remove the number of known categories and the number of scenes in known categories at 20\%, 50\%, and 80\% in the training set $D_{train}$, respectively. As shown in Table \ref{tab:4} and Table \ref{tab:5}, even if only 20\% of the categories in the training set can be assessed, our proposed RS-MetaNet still shows good results. For the UCMerced\_LandUse dataset, this means that only two categories can be used for training, but we need to make a judgement in five classification tasks when given only one label. Even with such harsh conditions, RS-MetaNet can still achieve classification accuracies of 43.53\% and 62.43\%, respectively. The same is true for Table \ref{tab:6} and Table \ref{tab:7}; RS-MetaNet still shows good robustness to the lack of training data.

	% =========
	% Fig. 05
	% =========
	
	% ===========
	% Table. 01
	% ===========
	\begin{table}[!t]
		\centering
		\caption{Classification accuracy (\%) of different meta training ratios of categories and given one label for testing on three datasets (average of 10 runs $\pm$ standard deviation).}
		\label{tab:4}
		\scalebox{0.85}{
			\begin{tabular}{cccc}
				\toprule
				& \multicolumn{3}{c}{Meta training ratios of categories} \\ \cline{2-4} 
				Data Sets         & 20\%            & 50\%            & 80\%            \\ \hline
				UCMerced\_LandUse & 43.53 $\pm$ 2.21    & 47.05 $\pm$ 1.99    & 53.18 $\pm$ 0.39    \\
				AID               & 45.93 $\pm$ 2.31    & 49.91 $\pm$ 1.19    & 53.63 $\pm$ 0.63     \\
				NWPU-RESISC45     & 44.56 $\pm$ 0.69    & 44.88 $\pm$ 0.86    & 45.78 $\pm$ 0.54    \\ \bottomrule
			\end{tabular}}
		\end{table}

		% ===========
		% Table. 02
		% ===========
		\begin{table}[!t]
			\centering
			\caption{Classification accuracy (\%) of different meta training ratios of categories and given five labels for testing on three datasets (average of 10 runs $\pm$ standard deviation).}
			\label{tab:5}
			\scalebox{0.85}{
				\begin{tabular}{cccc}
					\toprule
					& \multicolumn{3}{c}{Meta training ratios of categories}        \\ \cline{2-4} 
					Data Sets         & 20\%         & 50\%         & 80\%                \\ \hline
					UCMerced\_LandUse & 62.43 $\pm$ 1.87 & 65.40 $\pm$ 1.17 & 69.51 $\pm$ 0.26    \\
					AID               & 62.35 $\pm$ 1.79 & 66.54 $\pm$ 0.78 & 70.54 $\pm$ 0.68    \\
					NWPU-RESISC45     & 69.30 $\pm$ 0.87 & 70.91 $\pm$ 0.51 & 72.22 $\pm$ 0.33    \\ \bottomrule
				\end{tabular}}
			\end{table}
			
			% ===========
			% Table. 03
			% ===========
			
			\begin{table}[!t]
				\centering
				\caption{Classification accuracy (\%) of different meta training ratios of scenes per category and given one label for testing on three datasets (average of 10 runs $\pm$ standard deviation).}
				\label{tab:6}
				\scalebox{0.85}{
					\begin{tabular}{cccc}
						\toprule
						& \multicolumn{3}{c}{Meta training ratios of scenes per category}       \\ \cline{2-4} 
						Data Sets         & 20\%         & 50\%         & 80\%                 \\ \hline
						UCMerced\_LandUse & 44.18 $\pm$ 0.64  & 44.66 $\pm$ 0.39 & 52.57 $\pm$ 0.59    \\
						AID               & 40.77 $\pm$ 1.02  & 43.34 $\pm$ 0.52 & 50.55 $\pm$ 0.28    \\
						NWPU-RESISC45     & 40.98 $\pm$ 1.55  & 42.55 $\pm$ 1.08 & 44.51 $\pm$ 0.60    \\ \bottomrule
					\end{tabular}}
				\end{table}
				
				% ===========
				% Table. 04
				% ===========
				
				\begin{table}[!t]
					\centering
					\caption{Classification accuracy (\%) of different meta training ratios of scenes per category and given five labels for testing on three datasets (average of 10 runs $\pm$ standard deviation).}
					\label{tab:7}
					\scalebox{0.85}{
						\begin{tabular}{cccc}
							\toprule
							& \multicolumn{3}{c}{Meta training ratios of scenes per category}       \\ \cline{2-4} 
							Data Sets         & 20\%         & 50\%         & 80\%               \\ \hline
							UCMerced\_LandUse & 62.02 $\pm$ 0.87 & 64.34 $\pm$ 0.99 & 68.80 $\pm$ 0.62    \\
							AID               & 63.32 $\pm$ 0.56 & 65.43 $\pm$ 1.05 & 67.89 $\pm$ 0.55    \\
							NWPU-RESISC45     & 65.33 $\pm$ 0.99 & 68.78 $\pm$ 0.55 & 71.35 $\pm$ 0.36    \\ \bottomrule
						\end{tabular}}
					\end{table}
					
					% =========
					% Fig. 06
					% =========
					\begin{figure}[t]
						\begin{center}

							\subfigure[RS-MetaNet]{
								\includegraphics[width=3.8cm]{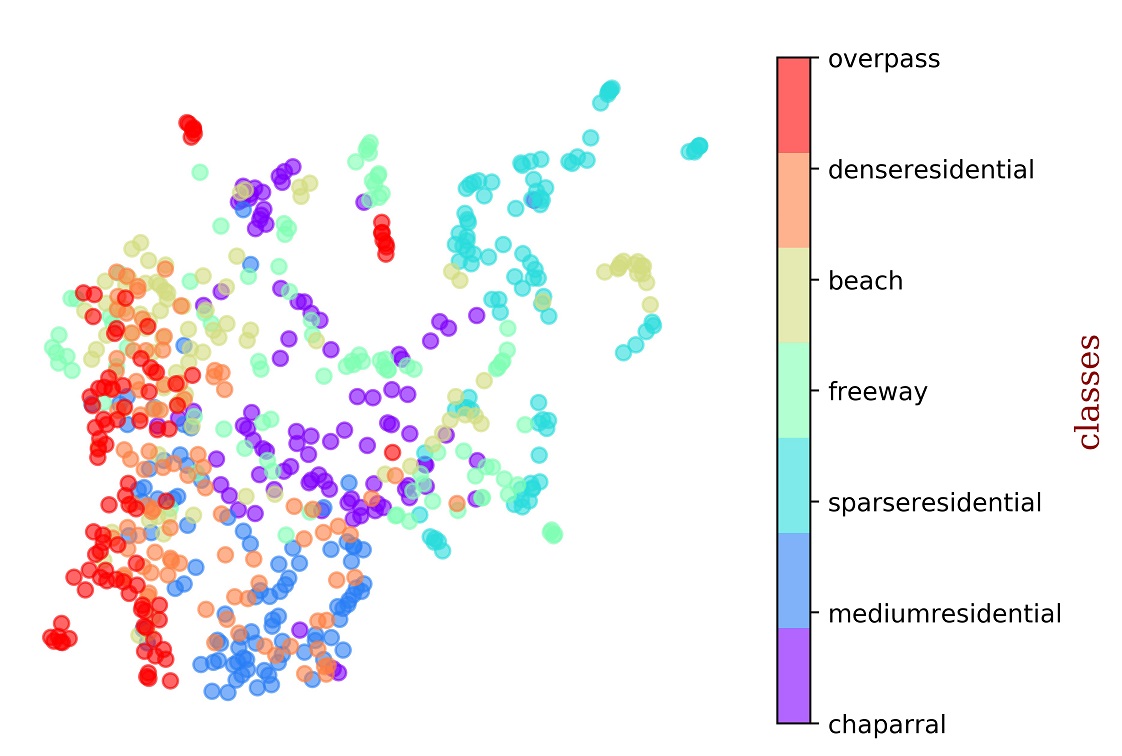}
							}
							\quad
							\subfigure[MAML]{
								\includegraphics[width=3.8cm]{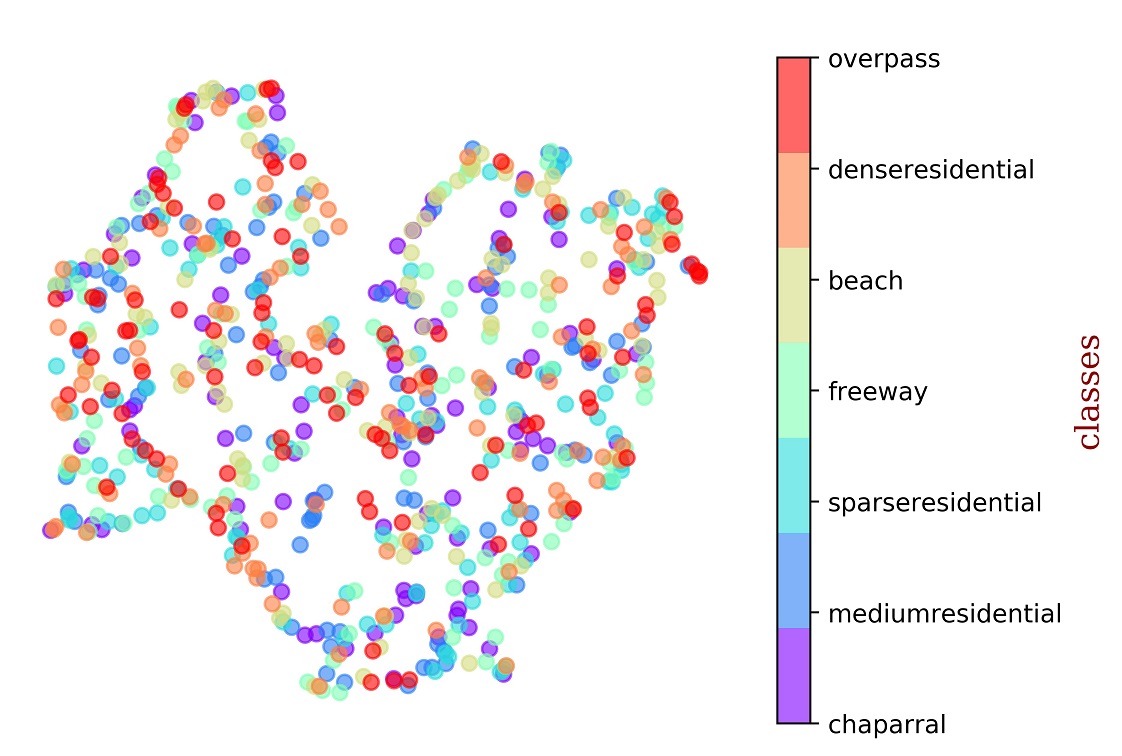}
							}
							
							\caption{Visualization results on the UCMerced\_LandUse dataset by UMAP when given only one label sample.}\label{umap}
							
						\end{center}
					\end{figure}
					
					% =========
					% Fig. 07
					% =========
					\begin{figure}[tbp]
						\begin{center}
							
							\subfigure[RS-MetaNet ]{
								\includegraphics[width=3.8cm]{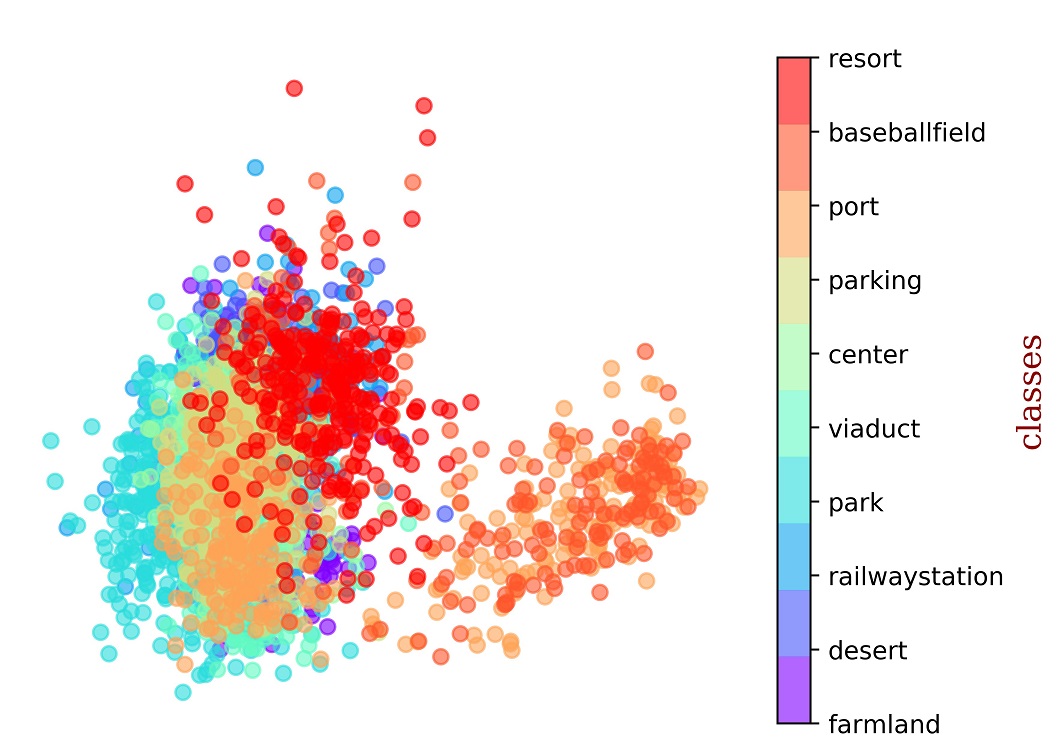}
							}
							\quad
							\subfigure[MAML ]{
								\includegraphics[width=3.8cm]{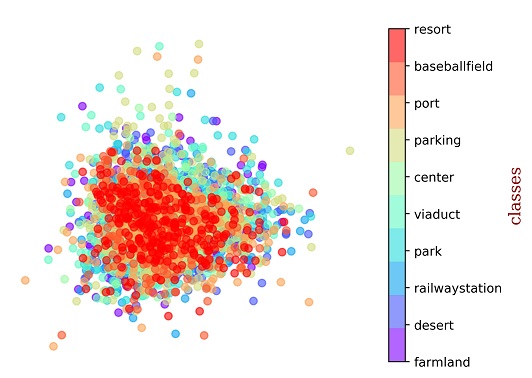}
							}
							\caption{Visualization results of RS-MetaNet (a) and MAML (b) on the AID by PCA with only one labelled sample. RS-MetaNet makes the space more distinctive.}\label{pca}
							
						\end{center}
					\end{figure}

					% =========
					% Fig. 08
					% =========
					\begin{figure}[t]
						\begin{center}

							\subfigure[RS-MetaNet ]{
								\includegraphics[width=3.8cm]{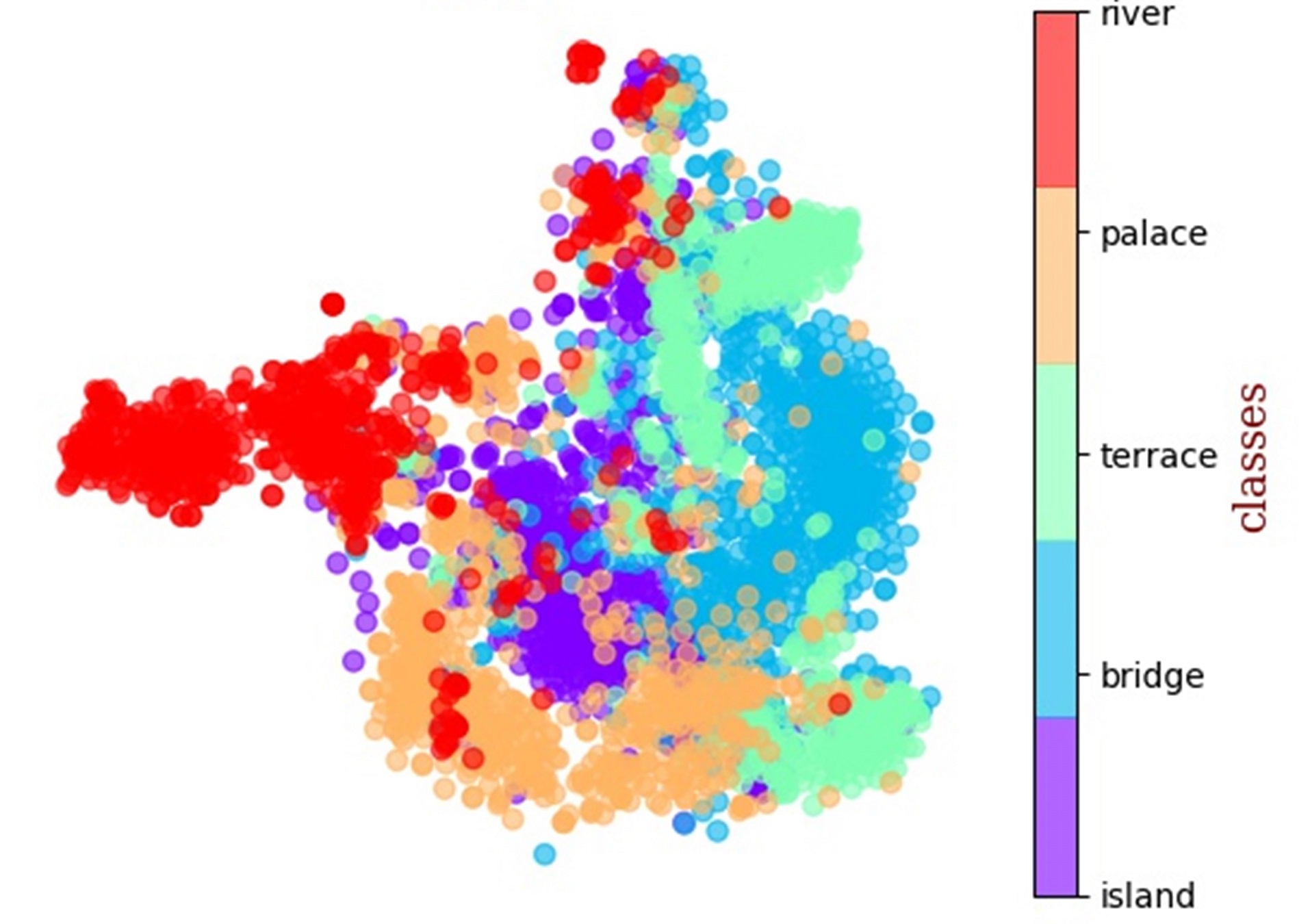}
							}
							\quad
							\subfigure[MAML ]{
								\includegraphics[width=3.8cm]{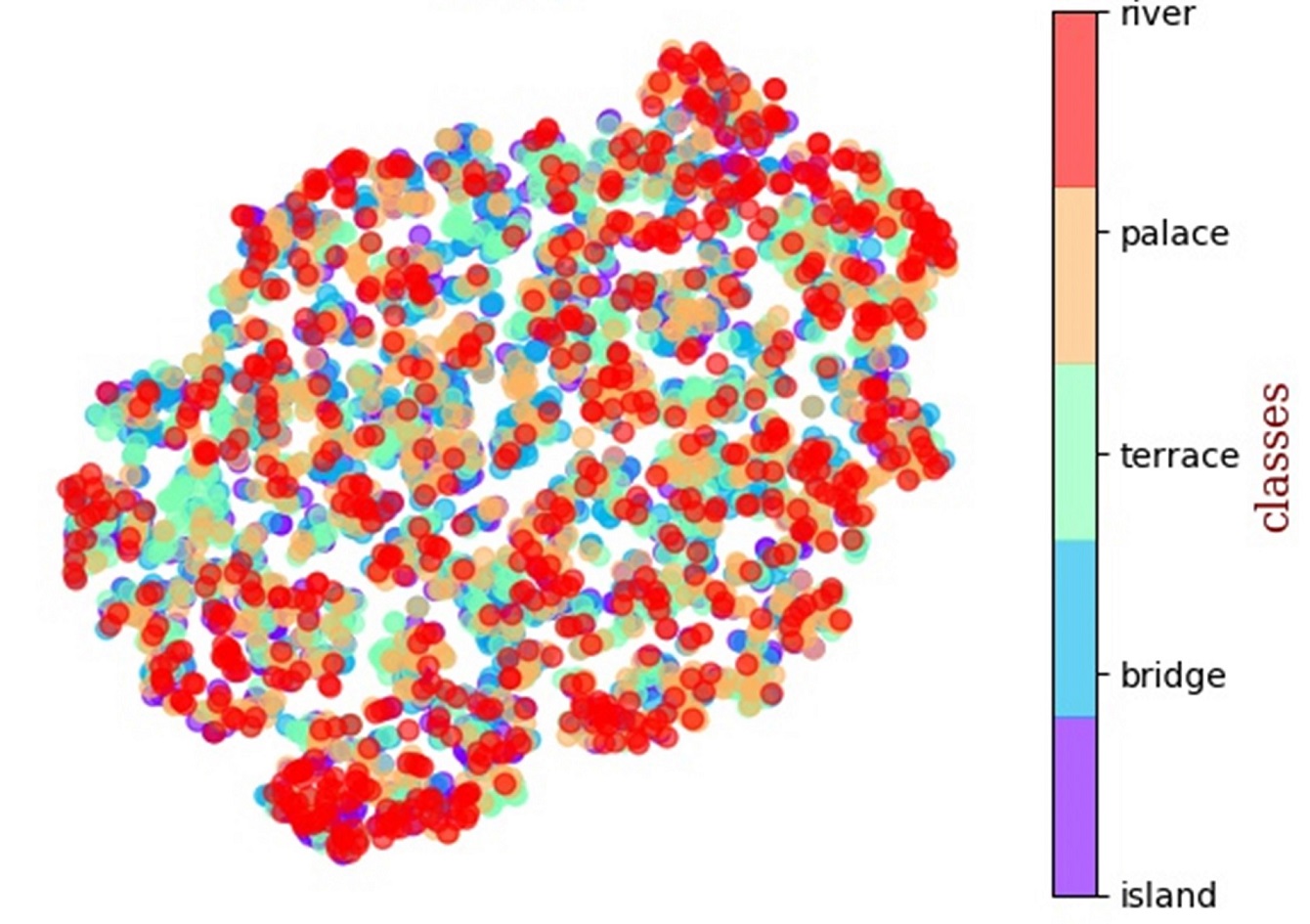}
							}
							\caption{Visualization results of RS-MetaNet (a) and MAML (b) on the NWPU-RESISC45 dataset by T-sne with only one labelled sample.}\label{tsne}
							
						\end{center}
					\end{figure}

					\subsection{Metric Space Analysis}\label{sec:Metric Space Analysis}
					
					As mentioned above and shown in Fig.\ref{overview}, our proposed model is designed to use only a small number of samples and learns to learn a metric space that can maximize the distance between different scenes, thereby making them more linearly separable. Therefore, when facing unknown remote sensing scenes with only a few samples, classification can be performed quickly and accurately. To illustrate this point, we provide only one labelled sample of each scene in the testing set and then input all the remaining data into two models. Thus, we do not use random sampling for verification here but rather all testing data. We then use Uniform Manifold Approximation and Projection (UMAP)\cite{RN60} to visualize the results on the UCMerced\_LandUse dataset and principal component analysis (PCA)\cite{RN52} to visualize the results on AID, as shown in Fig.\ref{umap} and Fig.\ref{pca}. The results indicate that our proposed RS-MetaNet achieves the desired effect very well and has better space discrimination than MAML, even though it has a heart shape.
					
					To further illustrate the effectiveness of our method and eliminate the model's preference for the descending dimension algorithm, we use another popular descending dimension visualization algorithm, T-sne\cite{RN53}, to visualize the effects of the model on the NWPU-RESISC45 dataset with one labelled sample. Because the NWPU-RESISC45 dataset has a richer amount of data, the results can be displayed more clearly. To make the results more distinct, we randomly sample five categories among 15 from $D_{test}$. As shown in Fig.\ref{tsne}, even if the number of data points increases, RS-MetaNet can still make the boundaries between different classes obvious, and there is a clear aggregation phenomenon among samples in the same class. In contrast, all the classes of the MAML algorithm are uniformly scattered throughout the space, and the boundaries between classes are blurred.

					Furthermore, Table \ref{tab:distance-ucm}$\sim$\ref{tab:distance-nwp} shows the model performance under different distance paradigms. The results demonstrate that compared to explicitly defining distances, such as cosine or Euclidean distances, our learnable metric allows us to make the most of the intrinsic information of the data and makes the model more effective.
					
					% ===========
					% Table. 05
					% ===========
					\begin{table}[t]
						\centering
						\caption{Classification accuracy (\%) of the different distance paradigms on the UCMerced\_LandUse dataset. L is the number of labelled samples for each class.
						}
						\label{tab:distance-ucm}
						\begin{tabular}{cccc}
							\toprule
							\multicolumn{1}{l}{}                   & \multicolumn{3}{c}{Number of labelled samples} \\ \cline{2-4}
							\multicolumn{1}{l}{Distance Paradigms} & L = 1      & L = 5     & L = 10     \\ \hline
							Cosine distance                        & 52.25      & 69.54     & 72.02      \\
							Euclidean distance                     & 53.24      & 71.19     & 72.56      \\
							Ours                                   & 55.29      & 71.42     & 75.16      \\ \bottomrule
						\end{tabular}
					\end{table}

					% ===========
					% Table. 06
					% ===========
					
					\begin{table}[!h]
						\centering
						\caption{Classification accuracy (\%) of the different distance paradigms on AID. L is the number of labelled samples for each class.
						}
						\label{tab:distance-aid}
						\begin{tabular}{cccc}
							\toprule
							\multicolumn{1}{l}{}                   & \multicolumn{3}{c}{Number of labelled samples} \\ \cline{2-4}
							\multicolumn{1}{l}{Distance Paradigms} & L = 1      & L = 5     & L = 10     \\ \hline
							Cosine distance                        & 50.55      & 68.2      & 76.35       \\
							Euclidean distance                     & 51.87      & 69.58     & 77.91       \\
							Ours                                   & 53.34      & 71.9      & 79          \\  \bottomrule
						\end{tabular}
					\end{table}
					
					% ===========
					% Table. 07
					% ===========
					\begin{table}[!h]
						\centering
						\caption{Classification accuracy (\%) of the different distance paradigms on the NWPU-RESISC45 dataset. L is the number of labelled samples for each class.
						}
						\label{tab:distance-nwp}
						\begin{tabular}{cccc}
							\toprule
							\multicolumn{1}{l}{}                   & \multicolumn{3}{c}{Number of labelled samples} \\ \cline{2-4}
							\multicolumn{1}{l}{Distance Paradigms} & L = 1      & L = 5     & L = 10     \\ \hline
							Cosine distance                        & 44.68                     & 64.76                     & 70.22                      \\
							Euclidean distance                     & 45.55                     & 65.35                     & 71.25                      \\
							Ours                                   & 46.22                     & 67.61                     & 73.16                      \\  \bottomrule
						\end{tabular}
					\end{table}

					\subsection{Effect of the Embedding Network Architecture}\label{sec:Effect Study}
					
					Table \ref{tab:acc-ucm}$\sim$\ref{tab:acc-nwp} shows the results when we vary the backbone among four different embedding architectures. When we use a standard four-layer convolutional network with a lower feature dimension, our model shows relatively good results. As the embedding dimension increases, the performance of the model improves further. Given only one labelled sample, using Googlenet as the embedding model, our RS-MetaNet achieves 56.99\%, 55.88\% and 50.12\% accuracy on the UCMerced\_LandUse, AID, and NWPU-RESISC45 datasets respectively. Using Resnet50 as the embedding model, our RS-MetaNet achieves 57.23\%, 52.78\% and 56.32\% accuracy on the three datasets, respectively. As the number of given labels increases, this boost effect is maintained. However, if the backbone is too complex, the model performance degrades due to meta-overfitting; that is, it constrains the hypothesis space of parameters too tightly around solutions to the source tasks. Overall, our method is flexible to substitution of the embedding model, and the networks with strong feature extraction capabilities have a positive impact on the model. The flexibility of the embedding model allows robustness to noisy embeddings and improves generalization.

					% =========
					% Table. 08
					% =========
					
					\begin{table*}[htbp]
						\centering
						\caption{Classification accuracy (\%) of the different methods for the UCMerced\_LandUse data set (average of 20 runs $\pm$ standard deviation; L is the number of labelled samples for each class, and the bold values represent the best accuracy among these methods in each case.)
						}
						\label{tab:acc-ucm}
						%\resizebox{0.8\textwidth}{!}{
						%\scalebox{1.25}{
						\begin{tabular}{clccc}
							
							\toprule
							\multicolumn{2}{l}{}                                            & \multicolumn{3}{c}{Number of labelled samples}                                    \\ \cline{3-5} 
							\multicolumn{1}{c}{Model}    & \multicolumn{1}{l}{Backbone}        & \multicolumn{1}{c}{L = 1} & \multicolumn{1}{c}{L = 5} & \multicolumn{1}{c}{L = 10} \\ \hline
							\multirow{4}{*}{Transfer learning based} & AlexNet   & 20.19 $\pm$ 0.78         & 25.08 $\pm$ 0.57         & 30.00 $\pm$ 0.68          \\
							& GoogleNet & 23.45 $\pm$ 1.29         & 45.22 $\pm$ 1.48         & 55.59 $\pm$ 0.94          \\
							& Resnet50  & 20.95 $\pm$ 1.16         & 29.23 $\pm$ 1.42         & 46.71 $\pm$ 1.45          \\
							& Resnet152 & 20.72 $\pm$ 0.41         & 29.29 $\pm$ 0.41         & 31.61 $\pm$ 1.55          \\ \hline
							\multirow{2}{*}{Meta learning based}     & Life long learning   & 39.47                 & 57.4                  & -                      \\
							& MAML                 & 47.53 $\pm$ 0.71         & 63.13 $\pm$ 0.92         & 64.99 $\pm$ 0.91          \\ \hline
							\multirow{4}{*}{Ours}                    & 4-layer-CNN              & 55.29 $\pm$ 0.59         & 71.42 $\pm$ 0.31         & 75.16 $\pm$ 0.29          \\
							& GoogleNet        & 56.99 $\pm$ 0.39         & 75.63 $\pm$ 0.04         & 80.65 $\pm$ 0.30         \\
							& Resnet50         & \textbf{57.23 \bm{$\pm$} 0.56}         & \textbf{76.08 \bm{$\pm$} 0.28}         & \textbf{81.23 \bm{$\pm$} 0.45}          \\
							& Resnet152        & 56.01 $\pm$ 0.91         & 72.68 $\pm$ 0.43         & 78.88 $\pm$ 0.11          \\    \bottomrule
						\end{tabular}
					\end{table*}

					% =========
					% Table. 09
					% =========
					\begin{table*}[!htbp]
						\centering
						\caption{Classification accuracy (\%) of the different methods for AID (average of 20 runs $\pm$ standard deviation; L is the number of labelled samples for each class; the bold values represent the best accuracy among these methods in each case.)}
						\label{tab:acc-aid}
						%\resizebox{0.8\textwidth}{!}{
						%\scalebox{1.25}{
						\begin{tabular}{cllll}
							\toprule
							\multicolumn{2}{l}{}                                            & \multicolumn{3}{c}{Number of labelled samples}                                    \\ \cline{3-5} 
							\multicolumn{1}{c}{Model}    & \multicolumn{1}{l}{Backbone}        & \multicolumn{1}{c}{L = 1} & \multicolumn{1}{c}{L = 5} & \multicolumn{1}{c}{L = 10} \\ \hline
							\multirow{4}{*}{Transfer learning based} & AlexNet   & 20.12 $\pm$ 0.97         & 24.56 $\pm$ 0.66         & 29.54 $\pm$ 0.62          \\
							& GoogleNet & 20.76 $\pm$ 1.29         & 40.67 $\pm$ 1.18         & 55.63 $\pm$ 1.35          \\
							& Resnet50  & 20.07 $\pm$ 1.55         & 29.61 $\pm$ 1.13         & 45.96 $\pm$ 1.32          \\
							& Resnet152 & 20.67 $\pm$ 1.09         & 23.66 $\pm$ 1.30         & 34.90 $\pm$ 1.62          \\ \hline
							Meta learning based                      & MAML                 & 47.93 $\pm$ 0.72         & 61.79 $\pm$ 0.75         & 69.90 $\pm$ 0.70          \\ \hline
							\multirow{4}{*}{Ours}                    & 4-layer-CNN              & 53.34 $\pm$ 0.16         & 71.90 $\pm$ 0.07         & 79.00 $\pm$ 0.45          \\
							& GoogleNet        & 55.88 $\pm$ 0.37         & 73.99 $\pm$ 0.05         & 79.85 $\pm$ 0.52          \\
							& Resnet50         & \textbf{56.32 \bm{$\pm$} 0.55}         & \textbf{74.48 \bm{$\pm$} 1.11}        & \textbf{80.57 \bm{$\pm$} 0.61}          \\
							& Resnet152        & 54.79 $\pm$ 0.48         & 73.23 $\pm$ 0.10         & 78.25 $\pm$ 0.67         \\ \bottomrule
						\end{tabular}
					\end{table*}

					% =========
					% Table. 10
					% =========
					\begin{table*}[!htbp]
						\centering
						\caption{classification accuracy (\%) of the different methods for the NWPU-RESISC45 dataset (average of 20 runs $\pm$ standard deviation; L is the number of labelled samples for each class; the bold values represent the best accuracy among these methods in each case.)}
						\label{tab:acc-nwp}
						%\resizebox{0.8\textwidth}{!}{
						%\scalebox{1.25}{
						\begin{tabular}{cllll}
							\toprule
							\multicolumn{2}{l}{}                                            & \multicolumn{3}{c}{Number of labelled samples}                                           \\ \cline{3-5} 
							\multicolumn{1}{c}{Model}    & \multicolumn{1}{l}{Backbone}        & \multicolumn{1}{c}{L = 1} & \multicolumn{1}{c}{L = 5} & \multicolumn{1}{c}{L = 10} \\ \hline
							\multirow{4}{*}{Transfer learning based} & AlexNet   & 20.02 $\pm$ 0.33            & 25.49 $\pm$ 0.52             & 29.99 $\pm$ 0.40          \\
							& GoogleNet & 25.91 $\pm$ 1.42            & 37.66 $\pm$ 1.37             & 55.35 $\pm$ 1.39          \\
							& Resnet50  & 21.14 $\pm$ 1.01            & 29.52 $\pm$ 1.16             & 49.62 $\pm$ 1.14          \\
							& Resnet152 & 20.71 $\pm$ 1.80            & 27.18 $\pm$ 1.56             & 35.18 $\pm$ 1.99          \\ \hline
							\multirow{2}{*}{Meta learning based}     & Life long learning   & \multicolumn{1}{c}{57.1} & \multicolumn{1}{c}{70.65} & \multicolumn{1}{c}{-}  \\
							& MAML                 & 42.29 $\pm$ 0.76            & 61.846 $\pm$ 0.81            & 68.77 $\pm$ 0.69          \\ \hline
							\multirow{4}{*}{Ours}                    & 4-layer-CNN              & 46.22 $\pm$ 0.41            & 67.61 $\pm$ 0.54             & 73.16 $\pm$ 0.44          \\
							& GoogleNet        & 50.12 $\pm$ 0.37            & 69.68 $\pm$ 0.14             & 76.28 $\pm$ 0.97         \\
							& Resnet50         &\textbf{52.78 \bm{$\pm$} 0.09}            & \textbf{71.49 \bm{$\pm$} 0.81}             & \textbf{77.37 \bm{$\pm$} 0.77}         \\
							& Resnet152        & 47.55 $\pm$ 0.66            & 68.47 $\pm$ 0.82             & 74.55 $\pm$ 0.96         \\ \bottomrule
						\end{tabular}
					\end{table*}

					\subsection{Ablation Study}\label{sec:Ablation Study}
					
\begin{figure}[tbp]
						\begin{center}
							
							\subfigure[Ablation of the meta training approach given one label]{
								\includegraphics[width=3.9cm]{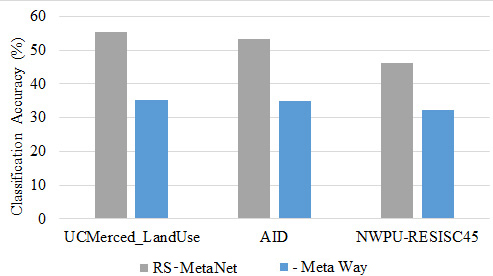}
							}
							\quad
							\subfigure[Ablation of the metric module given one label]{
								\includegraphics[width=3.9cm]{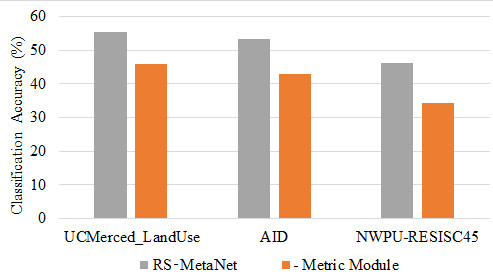}
							}
							\quad
							\subfigure[Ablation of the meta training approach given five labels]{
								\includegraphics[width=3.9cm]{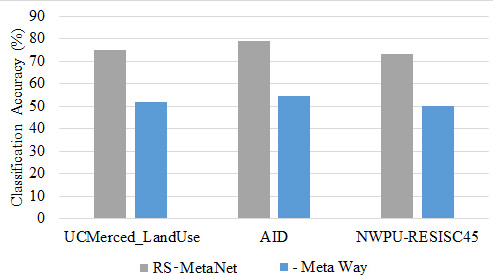}
							}
							\quad
							\subfigure[Ablation of the metric module given five labels]{
								\includegraphics[width=3.9cm]{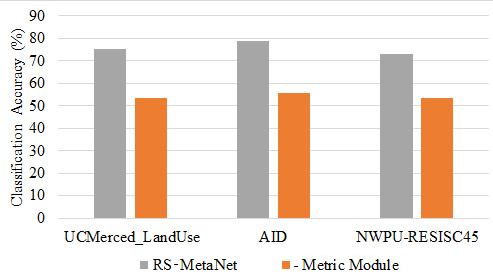}
							}
							\caption{\textbf{Ablation study} with the meta approach and metric module removed when given only one label and five labels separately.}.
							\label{ablation}
						\end{center}
					\end{figure}

					In this section, we will explore the meta training approach and the metric module on the overall framework. As shown in the left part of Fig.\ref{ablation}, when we ablate the meta training approach, the remote sensing scene classification accuracy is significantly reduced, regardless of whether one sample or five samples are given. This is because when the meta training approach is ablated, the model does not learn a task-based metric space but rather a traditional data-based metric space. Therefore, it will be susceptible to overfitting, as are other methods. The right side of Fig.\ref{ablation} is the result after ablating the metric module. Clearly, the accuracy is severely reduced without the metric module. This is because when the metric module is ablated, the model focuses more on the fitting of each task and ignores the spatial discriminability, which greatly reduces the generalization ability of the model. In addition, we found that the larger the number of labelled samples, the greater the role of the metric module in the model. For example, when there is only one labelled sample, the accuracy decreases by 6.45\% when the metric module is ablated, and when 10 samples are given, the accuracy decreases by 23.17\%, which means that the performance of the task-based metric space will improve as the number of labelled samples increases.

					% =========
					% Fig. 08
					% =========

					% =========
					% Fig. 09
					% =========
					
					\begin{figure}
						\begin{center}
							\includegraphics[width=3in]{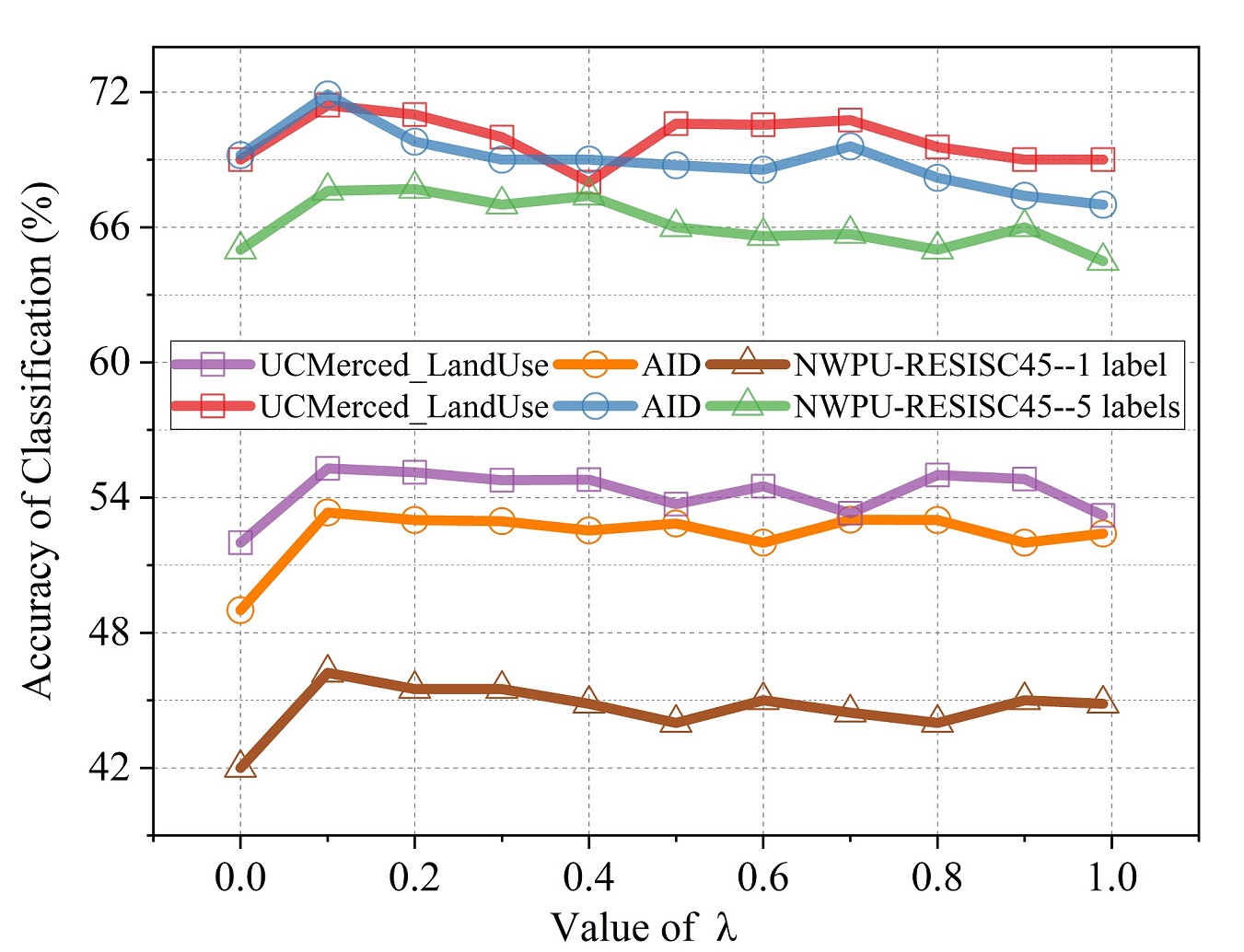}
							\caption{\textbf{Hyperparametric analysis.} The horizontal axis represents the different values of $\lambda$. The upper part of the picture shows the result when five labels were given, and the bottom shows the result when only one label was given.}
							\label{lambda}
						\end{center}
					\end{figure}

					\subsection{Hyperparameter Analysis}\label{sec:Hyperparameter Analysis}
					
					We also analysed the influence of an important hyperparameter $\lambda$, which controls the tilt for fitting and generalization and demonstrates the robustness of the proposed RS-MetaNet method. We performed a parametric analysis on all three datasets by taking different values of $\lambda$ in steps of 0.1 in the interval [0,1]. Fig.\ref{lambda} shows the results when only one labelled sample (bottom) and five labelled (top) samples are given. It can be seen from the figure that the model achieves the best effect when $\lambda$ is approximately 0.1; the classification accuracy is significantly higher than the accuracy when $\lambda = 0$, which shows that when maximizing the generalization ability, it is beneficial  to consider the fitting ability of the model. When $\lambda > 0.1$, as $\lambda$ increases, the model tends to fit an increasing amount of data, so the overall performance of the model shows a downward trend, which is consistent with our original goal of maximizing the generalization ability of the model. In addition, when the number of provided samples increases, the influence of $\lambda$ on the model decreases to some extent, but the overall trend does not change.

					\section{Discussion}\label{sec:Discussion}
					
\begin{figure}[tbp]
						\begin{center}
							
							\subfigure[split1]{
								\includegraphics[width=5cm]{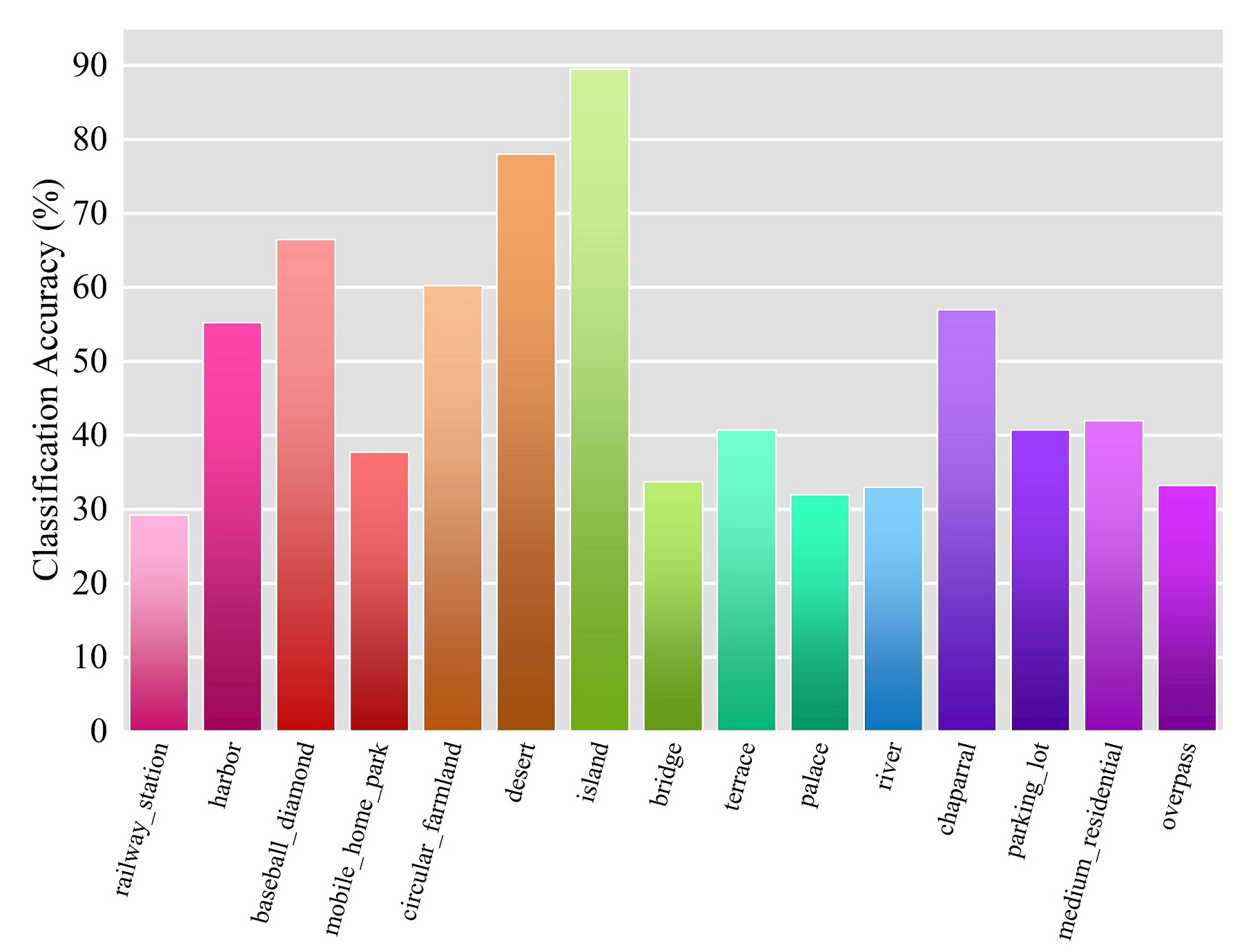}
							}
							\quad
							\subfigure[split2]{
								\includegraphics[width=5cm]{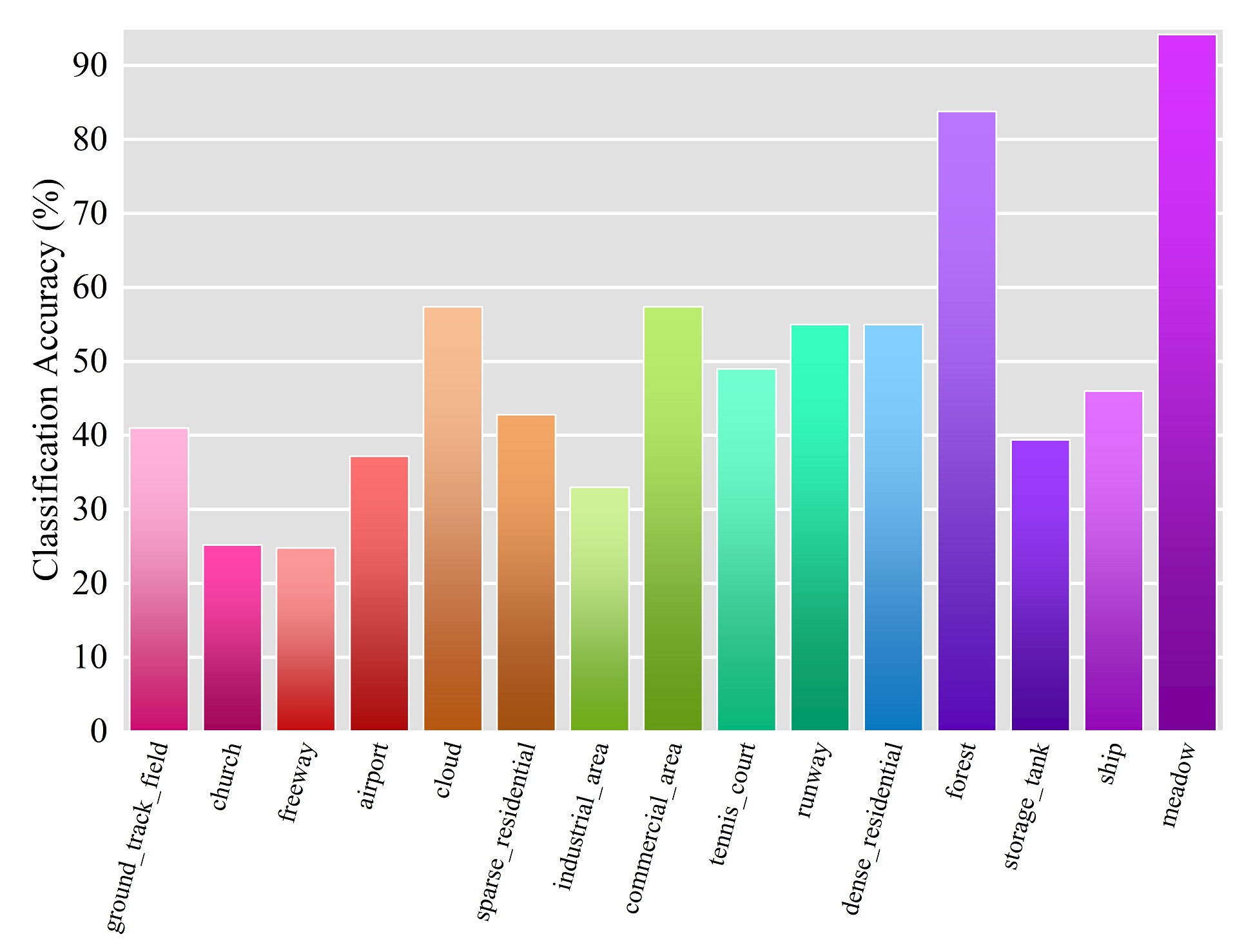}
							}
							\quad
							\subfigure[split3]{
								\includegraphics[width=5cm]{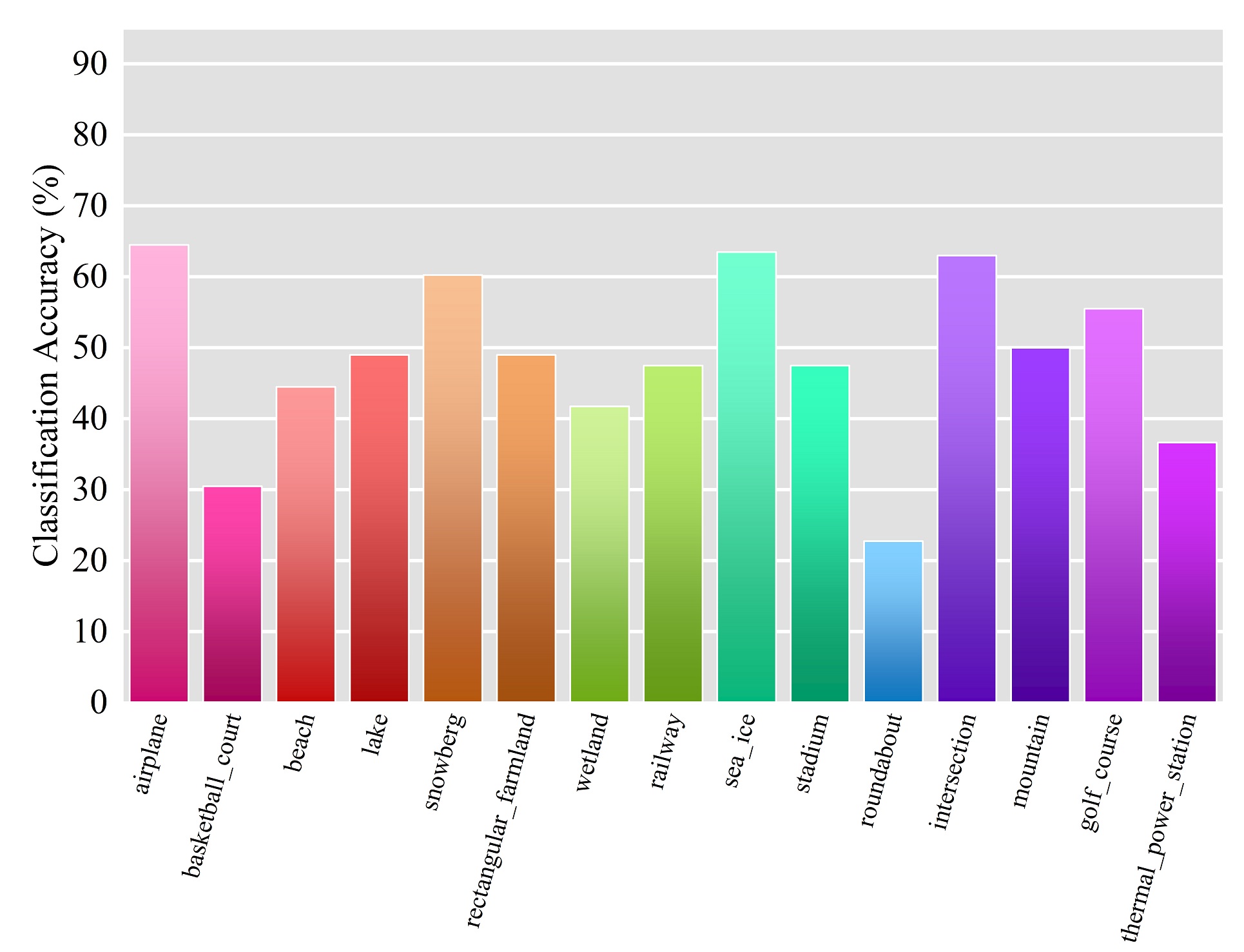}
							}
							\caption{The classification accuracy(\%) on different splits of the NWPU-RESISC45 dataset. Each split is alternately used as unseen classes for evaluation, with the other two splits as seen classes for training.}\label{discussion}
							
						\end{center}
					\end{figure}

					Our proposed RS-MetaNet method aims to learn a metric rule to make different remote sensing scenes more distinguishable in the space. We find that the performance on different remote sensing scenes shows a large gap. We divided the 45 scenes of the NWPU-RESISC45 dataset into three parts such that each part contained all the data of 15 scenes. We used two parts for training each time, and the remaining scenes were used to simulate few-shot remote sensing scene classification. The results are shown in Fig.\ref{discussion}. The classification results of different scenes are quite different, ranging from 30\% to 90\%. This shows that the discrimination of some scenes in the metric space could be improved further. Therefore, finding a way to selectively restrict these 'hard samples', such as by changing the class structure conversion strategy or using selective sampling rules, is an important goal for future work.

					% =========
					% Fig. 10
					% =========

					\section{Conclusion and Future Work}\label{sec:conclusion}
					This paper proposes a simple and effective framework, RS-MetaNet, for few-shot remote sensing scene classification in the real world. RS-MetaNet organizes training through a meta approach so that the learning level rises from data to tasks, which forces our model to learn task-based metrics. Task-based metrics learn task-level distributions that can be better generalized to unseen test tasks. Furthermore, we propose a new loss function, called Balance Loss, which guides our RS-MetaNet to gain powerful generalization capabilities on new samples by maximizing the distance between different categories while ensuring model fit.
					
					The differences in performance on different remote sensing scenes and the meta-overfitting problem that still exists in the meta-training process are our next challenges to solve. We conducted a preliminary exploration of these challenges, including adding regularization terms at each meta-training stage and using strategic sampling to repeatedly train hard samples. Overall, our RS-MetaNet method provides an effective reference for few-shot remote sensing scene classification in the real world.

					\ifCLASSOPTIONcaptionsoff
					\newpage
					\fi

					
					
					%\begin{thebibliography}{1}
					
					
					\bibliographystyle{IEEEtran}
					\bibliography{IEEEabrv,Bibliography}

					\vfill
					

				\end{document}